%% file: main.tex
\pdfoutput=1

\documentclass[11pt]{article}

\usepackage[]{EMNLP2023}

\usepackage{times}
\usepackage{latexsym}
\usepackage{makecell}
\usepackage[T1]{fontenc}

\usepackage[utf8]{inputenc}

\usepackage{microtype}

\usepackage{hyperref}
\usepackage{booktabs}
\usepackage{graphicx}
\graphicspath{{./imgs/}}
\usepackage{footmisc}
\usepackage{microtype}

\usepackage{caption}
\usepackage{subcaption}
\usepackage{array, makecell} %

\usepackage{pifont}

\usepackage{tabularx,colortbl}

\usepackage{tikz}
\usepackage{collcell}

\usepackage{etoolbox}

\newtoggle{inTableHeader}
\toggletrue{inTableHeader}
\newcommand*{\StartTableHeader}{\global\toggletrue{inTableHeader}}%
\let\OldTabular\tabular%
\let\OldEndTabular\endtabular%
\renewenvironment{tabular}{\StartTableHeader\OldTabular}{\OldEndTabular\StartTableHeader}%

\newcommand*{\MinNumber}{-1.0}%
\newcommand*{\MidNumber}{0.0} %
\newcommand*{\MaxNumber}{1.0}%

\newcommand{\ApplyGradient}[1]{%
  \iftoggle{inTableHeader}{#1}{
    \ifdim #1 pt > \MidNumber pt
        \pgfmathsetmacro{\PercentColor}{max(min(100.0*(#1 - \MidNumber)/(\MaxNumber-\MidNumber),100.0),0.00)} %
        \hspace{-0.33em}\colorbox{yellow!\PercentColor!blue}{#1}
    \else
        \pgfmathsetmacro{\PercentColor}{max(min(100.0*(\MidNumber - #1)/(\MidNumber-\MinNumber),100.0),0.00)} %
        \hspace{-0.33em}\colorbox{blue!\PercentColor!blue}{#1}
    \fi
  }}
\newcolumntype{R}{>{\collectcell\ApplyGradient}c<{\endcollectcell}}

\usepackage{amsmath}
\usepackage[utf8]{inputenc} 
\usepackage[T1]{fontenc}    
\usepackage{hyperref}       
\usepackage{url}            
\usepackage{booktabs}       
\usepackage{amsfonts}       
\usepackage{nicefrac}       
\usepackage{microtype}      
\usepackage{xcolor}         
\usepackage{listings}
\usepackage{algorithm}
\usepackage{algorithmic}
\usepackage{graphicx}
\usepackage{multicol}
\usepackage{CJKutf8}
\usepackage{booktabs}
\usepackage{subcaption}
\usepackage[utf8]{inputenc}
\usepackage{titlesec}
\usepackage[normalem]{ulem}
\usepackage{xspace}

\usepackage{multirow}
\usepackage{graphicx}
\usepackage{subcaption}
\usepackage{pythonhighlight}
\usepackage[font=small]{caption}
\usepackage{amsthm}
\usepackage{tabularray}

\usepackage[nameinlink]{cleveref}
\crefformat{section}{\S#2#1#3} 
\crefname{algorithm}{Alg.}{Algs.}
\crefformat{subsection}{\S#2#1#3}
\Crefname{equation}{Eq.}{Eqs.}
\Crefname{figure}{Fig.}{Figs.}

\usepackage[colorinlistoftodos,prependcaption,textsize=tiny]{todonotes}

\newcommand{\model}{\textsc{PromptChart}}

\definecolor{ao(english)}{rgb}{0.0, 0.5, 0.0}

\usepackage{caption}
\usepackage{soul}

\usepackage{float}
\usepackage{array}

\usepackage{multirow}
\usepackage{hhline}

\title{Do LLMs Work on Charts? Designing Few-Shot Prompts for Chart Question Answering and Summarization
}

\author{
Xuan Long Do$^{1}$, Mohammad Hassanpour$^{2}$, Ahmed Masry$^{2}$, Parsa Kavehzadeh$^{2}$, \\ \textbf{Enamul Hoque}$^{2}$\textbf{,} \textbf{Shafiq Joty}$^{3,4}$\thanks{Work done when the author was on leave from NTU}\\
$^1$National University of Singapore, $^2$York University, Canada,\\
$^3$Nanyang Technological University, Singapore, $^4$Salesforce Research \\
\texttt{xuanlong.do@u.nus.edu, \{mhpour, parsaka, enamulh\}@yorku.ca,}\\
\texttt{srjoty@ntu.edu.sg, ahmed.elmasry24653@gmail.com}\\
}
\begin{document}
\maketitle

\begin{abstract}


A 
variety of tasks have been proposed recently to facilitate 
exploration and analysis of 
charts such as chart QA and summarization. The dominant paradigm to solve these tasks has been to fine-tune a pretrained model on the task data. However, this approach is not only expensive but also not generalizable to unseen tasks. On the other hand, large language models (LLMs) have shown impressive generalization capabilities to unseen tasks with zero- or few-shot prompting. However, their application to chart-related tasks is not trivial as these tasks typically involve considering not only the underlying data but also the visual features in the chart image. We propose \model, a multimodal few-shot prompting  framework with LLMs for chart-related applications. By analyzing the tasks carefully, we have come up with a set of prompting guidelines for each task to elicit the best few-shot performance from LLMs. We further propose a strategy to inject visual information into the prompts. Our experiments on three different chart-related information consumption tasks show that with properly designed prompts LLMs can excel on the benchmarks, achieving state-of-the-art.

\end{abstract}

\section{Introduction} \label{sec:intro}

Data visualizations such as bar charts and line charts are frequently used for analyzing data, to derive important insights and make informed decisions \cite{hoque2022chart}. However, comprehending important patterns and trends from charts and answering complex  questions about them can be cognitively demanding \cite{doi:10.1080/10691898.2017.1321974, doi:10.1080/03323315.2018.1440248}. To support users in analyzing charts, various downstream tasks have been proposed such as chart question answering or ChartQA \cite{masry-etal-2022-chartqa, opencqa,lee2022pix2struct} and chart summarization \cite{kantharaj-etal-2022-chart}.

To date, the dominant strategy to tackle these downstream tasks is to fine-tune a pre-trained language model or vision-language model on each task. While this strategy generally yields good performance, it requires a large amount of labeled data for each downstream task as well as computational resources to train. Furthermore, recent studies have shown that even after fine-tuning, these models still struggle with queries that involve logical and arithmetic reasoning \cite{liu2022matcha, masry-etal-2022-chartqa, opencqa,cheng2023chartreader}. 

Meanwhile, large language models (LLMs) have demonstrated impressive generalization capabilities to unseen tasks through \emph{in-context learning} and/or  \emph{instructional tuning}. With in-context learning, LLMs can perform a new task only by looking at few examples without making any gradient update on the task data \cite{gpt3,liu2023pre}, while instructional tuning aims to enhance the model's instruction following capability by explicitly finetuning an LLM on large amount of multi-task instructional data \cite{weifinetuned,instructgpt}. Such generalization ability of LLMs is game changing compared to the predominant finetuning paradigm with two key advantages. First, it eliminates the need to train the model on each downstream task, which is highly resource consuming for LLMs. Second, such ability makes LM-as-a-service possible \cite{LM-as-a-service}, powering wide range of real world applications. Thus, there are emergent studies focusing on few-shot prompting for vision-language tasks \cite{alayrac2022flamingo, zhou2022learning,najdenkoska2023meta}.

 \begin{figure*}[t!]
     \centering
        \includegraphics[width=1\linewidth, trim={0cm 0cm 0cm 0cm},clip]{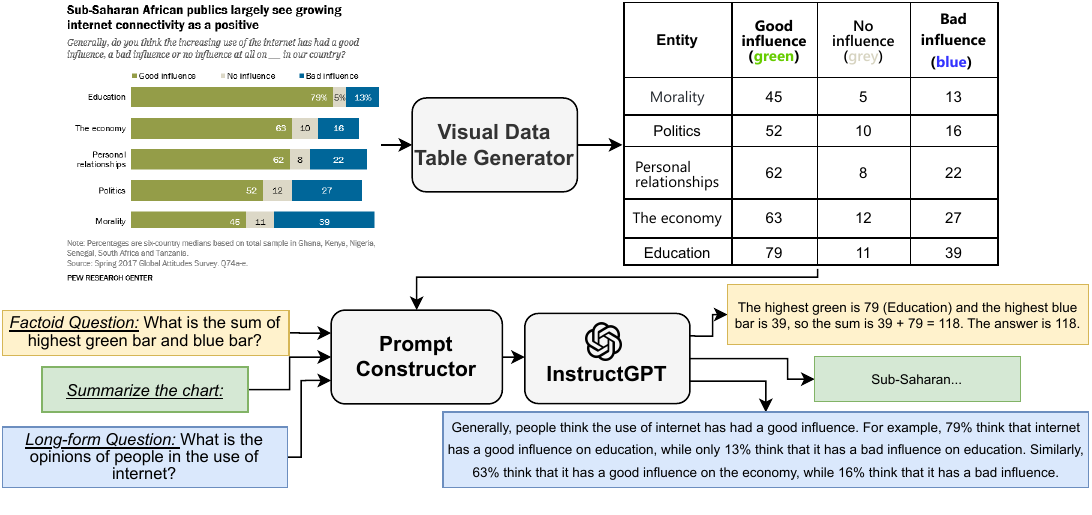}
         \caption{\small{An overview of our proposed \model\ prompting framework. The \emph{Visual Data Table Generator} module is omitted for the tasks of Long-form Chart Question Answering and Chart Summarization. 
         } 
         }
    \label{fiq:overview-framework}
     \vspace{-2mm}
\end{figure*}

Despite these recent advances, there has not been any comprehensive work on how few-shot prompting can be effectively applied to chart comprehension and reasoning tasks. \citet{liu2022deplot} present a one-shot method for ChartQA, which consists of two main steps: \emph{(i)} extracting the data table from an input chart; \emph{(ii)} constructing one-shot prompt to infer the output from an LLM. Nonetheless, this work only focuses on the one-shot setting and only one type of downstream task (i.e., factoid chart question answering). Moreover, it only uses the extracted data table to answer tasks without using visual features (e.g., colors, positions, and shapes of graphical marks) from the chart, which limits the model's ability to perform visual reasoning. To our knowledge, there has not been any work that explores few-shot prompting for a diverse range of chart comprehension and reasoning tasks.

To bridge this gap, we take the first step to study few-shot prompting for chart-related tasks. We design \model, a multimodal prompting framework consisting of three modules as shown in \Cref{fiq:overview-framework}. Our main contribution is the \emph{Prompt Constructor } module, which aims to construct effective few-shot prompts. For each task, we focus on analyzing carefully its subcategories and experiment with different few-shot prompting setups to verify the necessity and effectiveness of each demonstration. We then define a set of attributes for each task to guide the prompt constructions. To effectively encode the visual information into the few-shot prompts, we propose \emph{Visual Data Table Generator (VDTG)} module to generate the \emph{visual data table} - a data table that includes visual information such as colors and positions of the chart's labels. 

We evaluate \model\ on three different benchmarks: (1) Factoid QA with charts where the answer to a question is a short phrase; (2) Long-form Chart QA or LCQA where the answer is an explanatory text; (3) Chart summarization where the output is a summary of the chart. Compared with existing strong baselines, our framework achieves significant improvements in automatic metrics and human preferences across all downstream tasks, achieving state-of-the-art performance.

\section{Related Work}
Our work is mainly related to two lines of prior studies that we briefly describe here. An extended related work can be found in \Cref{addpx:extended-related-work}.

\subsection{Chart QA and Summarization} 
Recently, there has been growing interest in solving various chart-related downstream tasks. For example, the chart question answering (CQA) problem takes a question about a chart as input and produces the answer as output~\cite{hoque2022chart}. \citet{plotqa} and \citet{masry-etal-2022-chartqa} propose chart question answering benchmarks targeting factoid questions that require visual and arithmetic reasoning. Meanwhile, long-form chart question answering task (or open-ended CQA) \cite{opencqa} requires an explanatory answer for each question by reasoning over the chart image. In addition, chart summarization task \cite{kantharaj-etal-2022-chart} focuses on having a chart image as input and generating a natural language summary covering key insights from the chart. In this work, we verify the effectiveness of our proposed framework on the above three chart question answering and summarization tasks. We study these tasks because they require interactions between textual queries and chart images and also because there are adequate publicly available datasets for them. 


\subsection{Prompting} 
With the scaling of model sizes \cite{bert, GPT-2, gpt3, chowdhery2022palm}, large language models (LLMs) have demonstrated strong capabilities in solving multiple NLP downstream tasks only by conditioning on the input prompt which contains a few demonstrations (a.k.a., few-shot prompting). Such paradigms of prompting LLMs are called prompt-based learning or in-context learning \cite{liu2023pre,beltagy-etal-2022-zero}. This line of research has attracted great attention from the research community to solve downstream tasks including vision-language tasks \cite{alayrac2022flamingo, zhou2022learning,najdenkoska2023meta}.
However, exploring how chart-related downstream tasks could benefit from the above techniques has received limited attention. Recently, \citet{liu2022deplot} propose the first one-shot reasoning framework for chart question answering based on PaLM \cite{chowdhery2022palm}. Nonetheless, this framework suffers from several limitations as discussed in \Cref{sec:intro}. On the contrary, our study framework {\model} overcomes these limitations.

\section{Problem Formulation} \label{sec:problem-definition}

\paragraph{$\bullet$ Problem Definition} We focus on three chart-related tasks: factoid chart QA (FCQA), long-from chart QA (LCQA) and chart summarization (CS). A sample in FCQA can be expressed as $(C,D,Q,A)$, where $C$, $D$, $Q$ and $A$ stand for the corresponding chart image, data table, input question and answer, respectively. Similarly, a sample in LCQA can be expressed as $(C,T,O,Q,A)$, where $T$ and $O$ respectively represent the chart title and its OCR (optical character recognition) output; other symbols are similarly defined as in FCQA. Similarly, a sample in CS can be expressed as $(C,T,O,A)$, where $A$ represents the chart summary. The LLM can utilize all the data except $A$ as input, and is expected to produce $A$ as output.


\begin{table}[t]
  \centering
  \small
  \scalebox{.9}{
  \begin{tabular}{l|c|c}
  \toprule
    Datasets & \makecell{Type} & \makecell{\#Charts/\#QA pairs} \\
    \midrule
    ChartQA & Factoid CQA & 21.1K/32.7K\\
    OpenCQA & Long-form CQA & 7.7K/7.7K\\
    Chart-to-Text & Chart Summarization & 44K/44K\\
    
    \bottomrule
  \end{tabular}}
  \vspace{-3mm}
  \caption{\small{Statistics of the downstream benchmarks.}}
   \vspace{-3mm}
  \label{tab:dataset-summarization}
\end{table}

\paragraph{$\bullet$ Benchmarks} We choose ChartQA \cite{masry-etal-2022-chartqa} -- a benchmark which contains factoid chart question-answer pairs with visual and logical reasoning, OpenCQA \cite{opencqa} -- another QA benchmark where the answers are explanatory descriptions, Chart-to-Text \cite{kantharaj-etal-2022-chart} -- a benchmark for chart summarization, as the FCQA, LCQA, CS benchmarks respectively to study. For Chart-to-Text, it consists of two datasets, referred to as Pew and Statista. The statistics of the benchmarks are presented in \Cref{tab:dataset-summarization}.

\section{Methodology} \label{sec:method}
\Cref{fiq:overview-framework} shows our prompting framework \model, consisting of three modules: \emph{(i) Prompt Constructor (PC)}; \emph{(ii) Visual Data Table Generator (VDTG)}; \emph{(iii) InstructGPT} \cite{instructgpt}. In the case of FCQA, 
the chart image $C$ is provided to the VDTG module to generate the \emph{visual data table} $D_v$. With $D_v$ and $Q$ as input, the PC module then constructs the prompt, which is fed into InstructGPT to generate an answer $A$. For LCQA and CS tasks, the chart title $T$ and the OCRs $O$ are input to the PC module to construct the prompt without going through the VDTG module. We present the details of each module one by one below.

\subsection{Prompt Constructor (PC)} \label{subsubsec:prompt-engineering}

For each task, the prompt constructor module constructs the prompts to query the LLM. \citet{liu2022deplot} only use a one-shot prompt 
without conducting a detailed task-prompt analysis. We argue that task-prompt analysis is critical because it helps to identify the necessary demonstrations for each prompt to maximize the LLM's performance. To achieve this, we thoroughly analyze the task and propose a set of guidelines for prompt construction. 

\paragraph{$\bullet$}{\textbf{Factoid Chart Question Answering (FCQA)}} Inspired by Chain-of-Thought \cite{wei2022chain}, we propose \emph{chain of chart reasoning (CCR)}, a prompting strategy for FCQA. We categorize the task in FCQA into $9$ subcategories outlined in \Cref{tab:chartqa-task-analysis} and design a CCR prompt format specifically for each. Our CCR formats satisfy two main properties. Firstly, the formats describe clearly the reasoning process step by step, leading to the final answer for each query. 
Secondly, the elements in the CCR (operands, operators, visual attributes) need to be well-specified with clear relationships. Here, visual attributes include colors (e.g., `red', `green') and positions (e.g., `left', `right') of marks (e.g., bars and lines).
We follow these guidelines to carefully handcraft our prompt consisting of \textbf{6} demonstrations which are samples from ChartQA \cite{masry-etal-2022-chartqa}. We show an example below:

\noindent{\small {\textbf{Question:} \textit{Is the average value of Andean Latin America and Cambodia more than the value of Thailand?}}}

\noindent{\small {\textbf{CCR:}\emph{"\textcolor{magenta}{The value of Andean Latin America is 1.47 and the value of Cambodia is 0.77}. So \textcolor{blue}{the average value of Andean Latin America and Cambodia is ( 1.47 + 0.77 ) / 2 = 1.12}. \textcolor{magenta}{The value of Thailand is 0.39}. Since \textcolor{ao(english)}{1.12 > 0.39, the average value of Andean Latin America and Cambodia is more than the value of Thailand}. \textcolor{violet}{The answer is Yes}"}}}

We present all $6$ demonstrations in \Cref{tab:chartqa-demonstrations}, and discuss their details in \Cref{ssec:demonstration-selection-chartqa}. 


\begin{table} 
\centering
\scalebox{.44}{
\centering
\begin{tabular}
{p{1.5cm}|p{3cm}|p{11cm}}
\toprule
Tasks & Specifics - Supportive Case & Chain of Chart Reasoning (CCR) Prompt Format \\
\midrule
 & Visual Retrieval (Color \& Position) - Case 1 & \emph{No demonstration needed} \\
 
\cline{2-2} \cline{3-3} 
Retrieval & Numerical Retrieval - Case 2 & \emph{No demonstration needed} \\
\cline{2-2} \cline{3-3}
& Compositional Retrieval - Case 3 & \emph{No demonstration needed} \\
\cline{2-2} \cline{3-3} 
& Complex Retrieval - Case 9 & \texttt{\textcolor{olive}{operands with visual attributes}, \textcolor{magenta}{operands without visual attributes}, \textcolor{ao(english)}{reasoning}, \textcolor{violet}{result}.} \\

\midrule
 & Add \& Subtraction - Case 4 & \texttt{\textcolor{magenta}{operands without visual attributes}, \textcolor{blue}{operators}, \textcolor{ao(english)}{reasoning}, \textcolor{violet}{result}.}\\
\cline{2-2} \cline{3-3}
 & Division \& Multiplication - Case 5 & \texttt{\textcolor{magenta}{operands without visual attributes}, \textcolor{blue}{operators}, \textcolor{ao(english)}{reasoning}, \textcolor{violet}{result}.} \\
\cline{2-2} \cline{3-3} 
Reasoning & Visual Reasoning - Case 6 & \texttt{\textcolor{olive}{operands with visual attributes}, \textcolor{blue}{operators}, \textcolor{ao(english)}{reasoning}, \textcolor{violet}{result}.} \\\cline{2-2} \cline{3-3}
& Compositional Reasoning - Case 7 & \texttt{\textcolor{olive}{operands with visual attributes}, \textcolor{magenta}{operands without visual attributes}, \textcolor{blue}{operators}, \textcolor{ao(english)}{reasoning}, \textcolor{violet}{result}.}\\
\midrule
Boolean & All types of reasoning - Case 8 & \texttt{\textcolor{olive}{operands with visual attributes}, \textcolor{magenta}{operands without visual attributes}, \textcolor{blue}{operators}, \textcolor{ao(english)}{reasoning}, \textcolor{violet}{result}.}\\
\bottomrule
\end{tabular}
}
\caption{\small Task types and prompt format for factoid chart question answering. Supportive cases are illustrated in \Cref{fiq:ablation-prompt-chartqa}. Each element in the CCR is colored uniquely.}
\label{tab:chartqa-task-analysis}
\vspace{-5mm}
\end{table}

\paragraph{$\bullet$}{\textbf{Long-form Chart Question Answering (LCQA)}}
Similar to FCQA, our goal is to build a prompt consisting of demonstrations that represent most of the available long-form chart QA pairs. We adhere to the following three principles while designing the prompt for the LCQA task: (i) Descriptive answer to each question in the prompt should be focused and informative enough, i.e., should not be too long and beyond the scope of the question or too short and unable to capture the full extent of the answer. 
(ii) the prompt should cover the different QA pair types. (iii) If the LLM is inferred to generate an answer for a question that is in the prompt, that generated answer must be as close as the answer given in the prompt. This property ensures that the model does not hallucinate when answering the questions that it is instructed from the demonstrations, leading towards more factually-correct generations.


We carefully select our prompt having a total of \textbf{6} demonstrations from OpenCQA \cite{opencqa}. Among them, 2 questions are \emph{Describe and Summary}, i.e., questions asking to describe or summarize the chart based on statistical properties such as describing data distribution, outliers, and trends, 2 are \emph{Comparative} i.e., questions asking to compare the specified items in the chart, and 2 are \emph{Discover},  questions asking to derive key insights or findings from the chart. All the demonstrations are shown in \Cref{tab:opencqa-demonstrations} with details in \Cref{ssec:best-number-of-demonstration}.


\paragraph{$\bullet$}{\textbf{Chart Summarization (CS)}} 
Except design principle (i), the other two principles of LCQA are applicable to chart summarization. Specifically, the prompt should cover different summarization types (e.g., comparative, descriptive, reasoning) 
 and the LLM should not hallucinate on demonstrations. We select \textbf{3} demonstrations from the Pew dataset and \textbf{3} demonstrations for the Statista dataset of Chart-to-Text \cite{kantharaj-etal-2022-chart}. \Cref{tab:chart2text-demonstrations} presents all the demonstrations, which contains examples of three summary types: \emph{Perceptual and Cognitive} (e.g., cause and effect, trends), \emph{Statistical and Comparative} (e.g., min, max, higher, smaller), and \emph{Visual Encoding} (relating to visual information such as colors and positions).

\subsection{Visual Data Table Generator (VDTG)}

Prior work \cite{liu2022deplot} only considers the data table without any visual information from the chart, which makes the task of chart QA lose its visual nature. To study how the visual information affects the model performance in few-shot prompting, we introduce the VDTG module to construct \emph{visual data table}, a representation integrating the usual data table with visual attributes of the marks (\Cref{fiq:visual-data-table}). Given a chart image, the VDTG module is trained to generate a visual data table with two properties: (i) it has the associated colors (in natural language) of the rows/columns' labels; (ii) the rows and columns of the chart are sorted in their appearing positions in the chart. 

We construct the training dataset for VDTG module from the \emph{full version} of ChartQA at Github\footnote{https://github.com/vis-nlp/ChartQA}. To create the ground-truth visual data table for each chart image, we first create \emph{sorted data table} by collecting all the texts and coordinators of the bounding boxes of the chart and sorting them by the x-axis values and y-axis values. We also collect all the colors of the charts' data marks (e.g., bars and lines) by parsing their original SVG files. Since those colors are in hexadecimal format, we convert them into natural language colors by mapping them with the name of its closest color in \texttt{webcolors}\footnote{https://pypi.org/project/webcolors/} using a KDTree \cite{bentley1975multidimensional}. Still, we have noticed some hexadecimal codes are not present in the \texttt{webcolors} set, so we decided to manually check 300 random visual data tables and added 80 new mappings \texttt{\{hex, color name\}} to the set. Moreover, we remove the charts whose bounding boxes information is not fully provided. Overall, we obtain a total of 18,317 training samples and 1,056 validation samples. We fine-tune UniChart \cite{masry2023unichart} on our constructed dataset, a SoTA Chart-to-Table model, to generate the visual data tables. It is worth noting that this module is chart-type agnostic. Naturally, since it is trained on ChartQA, it supports the available chart types in ChartQA, including a variety of bar charts, line charts, and pie charts. The details of our training and inference for this module are 
in \Cref{appdx:vdtg-training-details}.


\subsection{InstructGPT} \label{subsubsec:instruct-gpt}

We use the frozen InstructGPT \cite{instructgpt} as our LLM, which is based on GPT-3 \cite{gpt3} and has been fine-tuned on natural human written instructions. Note that other LLMs such as ChatGPT
can also be directly used.


\section{Experiments} \label{sec:experimentation}

\subsection{QA and Summarization Experiments} 


\noindent{$\bullet$}{ \textbf{Baseline Models}} 
We compare our model with $3$ streams of baselines. The \emph{first stream} consists of \emph{Models with fine-tuning strategies}, which can be further categorized into two subcategories: (a) models that take the chart image as input such as Pix2Struct \cite{lee2022pix2struct} - the state-of-the-art visually-situated language model, MatCha \cite{liu2022matcha}, and UniChart \cite{masry2023unichart} - the state-of-the-art chart pretraining model; and (b) models that take the data table as input such as VisionTapas \cite{masry-etal-2022-chartqa} - the original state-of-the-art vision-language model on ChartQA, VL-T5 \cite{cho2021unifying} - the state-of-the-art model on OpenCQA, T5 \cite{2020t5} a competitive text generation model on the chart summarization task as reported by \citet{kantharaj-etal-2022-chart}. The \emph{second stream} consists of \emph{ Large language models with data tables provided}: FlanT5-xxl \cite{chung2022scaling} - an instructions-fine-tuned large language model achieving impressive few-shot results on many NLP tasks, PAL \cite{gao2022pal} - the state-of-the-art few-shot code-based reasoning model on the GSM8K benchmark \cite{cobbe2021gsm8k}. The \emph{third stream} comprises \emph{Large language models without data tables provided}: DePlot + FlanPaLM \cite{liu2022deplot} - the state-of-the-art factoid chart question answering model.

For the FCQA task, we select Pix2Struct, MatCha, DePlot, PAL, FlanT5-xxl as the five baselines. We adopt FlanT5-xxl and VL-T5 as the two baselines for the LCQA task. For the CS task, we compare our model with FlanT5-xxl and T5. 

\paragraph{$\bullet$}{ \textbf{Inference}}
InstructGPT model is called for the inference through OpenAI API being \emph{text-davinci-003} in April 2023. We utilize Nucleus Sampling \cite{Holtzman2020The} as our decoding strategy with a temperature value of 0.7, a max\_tokens value of 256, and a top\_p value of 0.9 as our decoding hyper-parameters. The frequency\_penalty and presence\_penalty are set to 0.

\paragraph{$\bullet$}{\textbf{Automatic Evaluation \& Human Preferences}}
For text generation tasks, prior works \cite{belz-reiter-2006-comparing,tan-etal-2015-awkward,liu2023gpteval} discuss that n-gram reference-based evaluation metrics for text generation tasks such as BLEU \cite{papineni2002bleu} and ROUGE \cite{lin-2004-rouge} may have a relatively low correlation with human judgments and generated texts with very high scores can have a poor quality \cite{smith-etal-2016-climbing, kann-etal-2018-sentence,mathur-etal-2020-tangled} 
. Therefore, we focus on evaluating the factual correctness of the models for LCQA and CS tasks. In particular, we use \emph{QAFactEval} \cite{fabbri2022qafacteval} (by averaging all its criteria), a QA-based metric that computes a factual score based on the ability of a QA model to answer questions generated from the input, given the generated text. We further follow \citet{instructgpt} to conduct \emph{human preferences} to evaluate the performance of models. Specifically, for each dataset, we randomly select $300$ samples. We hire three annotators who are English native speakers and collect the generated outputs of these samples from the best-performing baseline (which is not a variant of our model) and \model. We then ask the annotators to rate which one they prefer, or it is a tier. The annotators' agreements are measured by Kripp.'s alpha \cite{krippendorff2011computing}. 


\paragraph{ $\bullet$ Zero- and Few-shot Settings}

For zero-shot, we follow 
\citet{kojima2022large} to prompt the InstructGPT model to generate the textual outputs step by step. Specifically, we use \texttt{"Answer the following question step by step."} as the FCQA instruction and \texttt{"Answer the following question step by step by a single paragraph."} as the LCQA instruction. For the CS task, the instruction is \texttt{"Summarize the trends in the chart step by step and write the summary."}. The full sets of few-shot demonstrations for FCQA, LCQA, CS tasks are provided in \Cref{tab:chartqa-demonstrations}, \Cref{tab:opencqa-demonstrations}, \Cref{tab:chart2text-demonstrations} respectively.

\begin{table}[t!]
\centering
\setlength{\tabcolsep}{3pt}
\scalebox{0.65}{
\begin{tabular}{lcc|ccc}
\toprule
\multirow{2}{*}{Model} & \multicolumn{1}{c}{\textbf{Gold table?}} & \multicolumn{1}{c}{\textbf{Mode}} & \multicolumn{3}{c}{\textbf{ChartQA}} \\
 & & & Aug. & Human & Avg. \\
\midrule
T5 & Yes & fine-tuned & 56.96 & 25.12 & 41.04 \\
VLT5 & Yes &  fine-tuned & 56.88 & 26.24 & 41.56 \\
VisionTapas & Yes &  fine-tuned & 61.44 &  29.60 & 45.52 \\
Pix2Struct & No &  fine-tuned & 81.60 & 30.5 & 56.0 \\
MatCha & No & fine-tuned & \textbf{90.2} & 38.02 & 64.2 \\
UniChart & No & fine-tuned & 88.56 & 43.92 & 66.24 \\
\midrule
FlanT5-xxl & Yes & zero-shot & 63.84 & 27.84 & 45.84  \\
PAL & Yes & zero-shot & 80.88 & 42.77 & 61.83 \\
InstructGPT & Yes & zero-shot & 62.24 & 40.96 & 51.60 \\
DePlot + Inst. + CCR & No & zero-shot & 78.64 & 45.78 & 62.21 \\
UniChart + Inst. + CCR & No & zero-shot & 79.76 &  46.67 & 63.28  \\
\textbf{\model} & No & zero-shot & 79.44 & 47.11 & 63.1 \\
\midrule
DePlot + GPT-3 + CoT & No & one-shot &  37.3 &  36.5 & 36.9  \\
DePlot + FlanPaLM + CoT & No & one-shot & 76.7 & 57.8 & 67.3  \\
\midrule
FlanT5-xxl & Yes & few-shot & 64.56 & 28.56 & 46.56 \\
PAL & Yes & few-shot & 0.00 & 0.16 & 0.08  \\
InstructGPT & Yes & few-shot & 75.12 & 49.52 & 62.32 \\
DePlot + Inst. + CCR & No & few-shot & 80.28 & 60.49 & 70.39  \\
UniChart + Inst. + CCR & No & few-shot & 81.68 & 62.24 & 71.96 \\

\midrule
\textbf{\model} & No & few-shot & 81.44 & \textbf{63.2} & \textbf{72.32} \\
\bottomrule
\end{tabular}
}
\caption{\small {Factoid Chart QA experimental results. {\model} is equivalent to InstructGPT + CCR + Visual Information. All the metrics are measured in percentage.} 
}
\vspace{-3mm}
\label{tab:prompt-chart-chatqa-results}
\end{table}

\subsection{Main Results}

\paragraph{$\bullet$}{\textbf{VDTG Results}
To evaluate our visual data table generation model, we follow recent relevant works in the Chart-to-Table domain and evaluate our model using the Relative Number Set Similarity (RNSS) \cite{masry-etal-2022-chartqa} and Relative Mapping Similarity (RMS) \cite{liu2022deplot} metrics. Our model exhibits an impressive RNSS of 94.29\% and RMS of 89.55\%, highlighting its strong capability to extract data from images. This successful visual data table extraction process has significantly contributed to the overall performance of {\model}.

\paragraph{$\bullet$}{\textbf{FCQA, LCQA, CS Results} Our main experimental results for FCQA task on ChartQA are outlined in \Cref{tab:prompt-chart-chatqa-results}. We derive three main observations. Firstly, compared to zero-shot, one-shot, and fine-tuned baselines, {\model} (few-shot) outperforms them by large margins, especially on the \emph{human} test set. 
These improvements verify the effectiveness of our proposed FCQA prompt. Secondly, our experiments on \emph{augmented} test set reveal that the \model{} gains minor improvements when conditioning on the few-shot prompt instead of the instruction only (i.e., the zero-shot setting). We attribute this to the nature of the questions from \emph{augmented} test set since they are mostly retrieval and not reasoning questions,} 
which makes them less challenging for LLMs such as InstructGPT. Thirdly, the results of {\model} and its variant \emph{UniChart + InstructGPT + CCR} indicate that although the visual data tables slightly worsen the performance of the model on visually-irrelevant questions, they boost the performance of the models on visually-related questions, leading to overall improvements. We further provide additional analysis for the visual data table representation in \Cref{appdx:analysis-visual-information}. Exemplary cases are in \Cref{tab:chatqa-cases}. 

\begin{table}[t!]
\centering
\setlength{\tabcolsep}{3pt}
\scalebox{.45}{
\begin{tabular}{lc|cc|cccc}
\toprule
\multirow{2}{*}{Model} & \multicolumn{1}{c}{\textbf{Mode}} & \multicolumn{2}{c}{\textbf{OpenCQA}} & \multicolumn{4}{c}{\textbf{Chart-to-Text}}\\
 & & QAFactEval & HumP & QAFactEval-P & HumP-P & QAFactEval-S & HumP-S \\
\midrule
T5 & fine-tuned  & 21.39 & - & 18.57 & - & \textbf{53.69} & 2.34 \\
VLT5 &  fine-tuned & 22.65 & 10.34 & - & - & - & - \\
MatCha & fine-tuned & - & - & 13.03 & - & 35.50 & -  \\
\midrule
FlanT5-xxl & zero-shot & 13.97 & - & 19.36 & 4 & 23.98 & - \\
InstructGPT & zero-shot & 23.16 & - & 18.13 & - & 24.72 & - \\
\textbf{\model} & zero-shot & 25.85 & - & 18.44 & - & 25.31 & -  \\
\midrule
FlanT5-xxl & few-shot & 5.61 & - & 18.29 & - & 34.05 & - \\
InstructGPT & few-shot & 24.23 & - & 29.17 & - & 38.02 & - \\
\textbf{\model} & few-shot & \textbf{27.19} & \textbf{81.67} & \textbf{31.17} & \textbf{94.34} & 49.88 & \textbf{94.23} \\
\bottomrule
\end{tabular}
}
\caption{\small{Long-form chart question answering and chart summarization experimental results. Metrics are measured in percentage. 
}
}
\label{tab:prompt-chart-experimentation}
\vspace{-5mm}
\end{table}


Our evaluations in \Cref{tab:prompt-chart-experimentation} for LCQA and CS tasks on OpenCQA and Chart-to-Text benchmarks illustrate a number of noteworthy findings. Firstly, with the carefully-designed prompts that we propose, \model{} significantly improves the factuality scores on OpenCQA and Chart-to-Text Pew datasets, surpassing the performance of all fine-tuned baselines. Secondly, the model also exhibits superior human preferences in comparison to these fine-tuned methods. Finally, we observe that {\model} generally performs better in the few-shot setting compared to the zero-shot setting. This highlights the importance of good demonstrations in determining its performance. Our annotators achieve strong agreements with an average Kripp.’s alpha values of 0.86 (0.79 for OpenCQA, 0.91 for Chart-to-Text-Pew, and 0.88 for Chart-to-Text-Statista).

\section{Discussion} \label{sec:discussion}


\subsection{Demonstration Selection for FCQA} \label{ssec:demonstration-selection-chartqa}

\begin{figure*}[t!]
     \centering
        \includegraphics[width=1\textwidth,trim={0cm 0cm 0cm 0cm},clip]{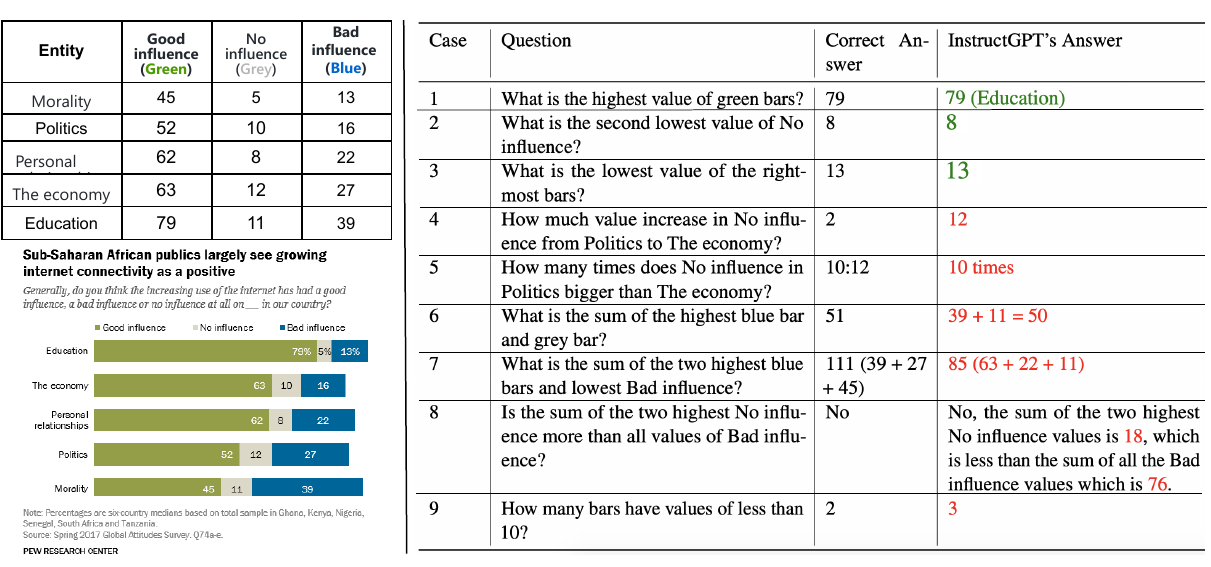}
         \vspace{-3mm}
        \caption{\small{Supportive cases for our ChartQA prompt construction. \textcolor{ao(english)}{Green} answers are correct answers, and \textcolor{red}{red} ones are errors.  
         }}
    \label{fiq:ablation-prompt-chartqa}
    \vspace{-3mm}
\end{figure*}

We discuss the necessity of each demonstration in our proposed six-shot prompt for FCQA task. \Cref{fiq:ablation-prompt-chartqa} presents our supportive cases and the InstructGPT's answers in the \emph{zero-shot} setting. By examining these cases, we derive several observations. Firstly, InstructGPT model can do \emph{Visual Retrieval}, \emph{Numerical Retrieval}, and \emph{Compositional Retrieval} without any demonstration (i.e., in zero-shot setting), which are proven by Case 1, 2, 3 respectively. Secondly, without any demonstration, the model seems to fluctuate in getting correct answers for arithmetic reasoning types including \emph{Add \& Subtraction} and \emph{Division} (Case 4, 5, 6, 7). 
Thirdly, for more complex reasoning types such as \emph{Compositional Reasoning}, the model performs poorly (Case 9) even in the reasoning steps (Case 8). Based on these observations, we propose the prompt consisting of $6$ demonstrations for the types of queries in which the model makes mistakes. We also conduct an experiment in which we select 6 random samples in the ChartQA validation set as demonstrations, \model{} achieves an overall accuracy of $68.92$\% (56.95\% in the human set, and 80.88\% in the augmented set), which is far lower from our proposed prompt, suggesting that our prompt is generalizable to other variations of questions within the benchmark.
Note that, for a new benchmark, our current selection of demonstrations may not be sufficient, however, our methodology of selecting demonstrations can be adapted to improve generalizability.

\subsection{Best Number of Demonstrations for LCQA and CS Task} \label{ssec:best-number-of-demonstration}




\begin{figure}[t!]
     \centering
     \includegraphics[width=.5\textwidth,trim={0cm 0cm 0cm 0cm},clip]{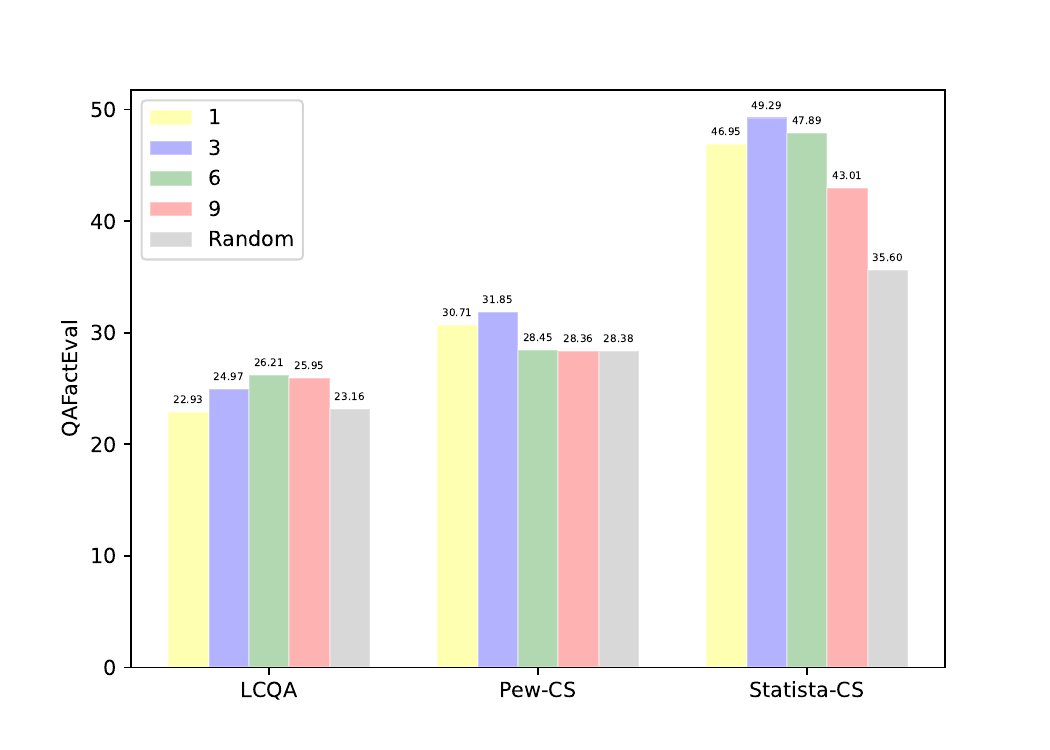}
     \vspace{-3mm}
     \caption{\small{Performance of our {\model} with different numbers of demonstrations in the prompt on 300 random samples of each benchmark.
     }
     }
    \label{fiq:ablation-number-of-demonstrations}
    \vspace{-4mm}
\end{figure}

We conduct experiments to select the best number of demonstrations in the prompts fed to the model for the tasks of LCQA and CS. Specifically, we select a subset containing 300 random samples from the validation splits of Chart-to-Text (Pew dataset and Statista dataset) and OpenCQA as our mini-test sets. We then compare the performance of our model at different numbers of carefully-selected demonstrations (1, 3, 6, 9) and report the corresponding QAFactEval scores \cite{fabbri2022qafacteval}. We select the demonstrations thoroughly, following the proposed properties in \Cref{sec:method}. After obtaining the best number of demonstrations on each mini-test set, we randomly select the same number of demonstrations to test the effectiveness of our proposed properties. The results are presented in \Cref{fiq:ablation-number-of-demonstrations}. We derive two main observations.Firstly, as we add more demonstrations to the prompt, the model may tend to generate more factually incorrect texts compared to using the previous number of demonstrations. 
Secondly, our careful sets of demonstrations boost the performance of the model significantly compared to random selections, which illustrates the effectiveness of our proposed properties.


\subsection{Dive Deeper into the Proposed Properties} \label{ssec:ablation-proposed-properties}

\begin{table}
\centering
\scalebox{.9}{
\begin{tabular}{lccc}
\midrule
Model & OpenCQA & Pew & Statista \\ 
\midrule
LCQA w/o \emph{(ii)} & 25.87 & - & - \\
LCQA w/o \emph{(iii)} & 27.01 & - & - \\
\midrule
\model & \textbf{27.19} & - & - \\
\midrule
\midrule
CS w/o \emph{(ii)} & - & 29.11 & 47.61 \\
CS w/o \emph{(iii)} & - & 30.79 & 48.16 \\
\midrule
\model & - & \textbf{31.17} & \textbf{49.88} \\
\midrule
\end{tabular}
}
\caption{Effects of proposed properties. We select the same number of demonstrations with our proposed number for each task as in \Cref{sec:method}.}
\vspace{-6mm}
\label{tab:effects-proposed-properties-table}
\end{table}

To verify the necessity of our proposed properties for guiding the prompt construction in \Cref{sec:method}, we conduct experiments by removing the properties one by one, resulting in $4$ 
experiments. Our illustrative cases are presented in \Cref{appdx:proposed-properties-illustrative}:

{$\bullet$} \textbf{LCQA - w/o prop. \emph{(ii)}} Our prompt for LCQA task only contains 6 \emph{Describe and Summary} question-answer pairs. This experiment is to show that without \emph{Comparative} and \emph{Discoverer} QA pairs in the prompt, the model achieved lower performance.

{$\bullet$} \textbf{LCQA - w/o prop. \emph{(iii)}} We select $6$ other samples in OpenCQA dataset as demonstrations such that the model hallucinates on demonstrations, showing that it is better to ensure this property.

{$\bullet$} \textbf{CS - w/o prop. \emph{(ii)}} Our prompt in the CS task only contains $6$ \emph{Statistical and Comparative} summaries. This experiment is to verify that without any \emph{Perceptual and Cognitive} summary in the demonstrations, the model might output numbers or texts that do not exist in the input.}

{$\bullet$} \textbf{CS - w/o prop. \emph{(iii)}} We select $6$ other summaries in the dataset as demonstrations such that there exists at least one of them which the model hallucinates on, showing that it is better to ensure property \emph{(iii)}.

Our experimental results are outlined in \Cref{tab:effects-proposed-properties-table}. 
We observe that removing any proposed property results in decreased performance of the model, which enforces our proposed properties.  

\subsection{Error Analysis} \label{sec:error-analysis}
Despite achieving promising performance, our proposed framework still falls into several categories of errors. We discuss them below and all the case studies are introduced in \Cref{tab:error-analysis}.

\paragraph{$\bullet$}{\textbf{Factoid Chart Question Answering}} To understand the challenges of FCQA task and promising future directions, we manually investigate 100 FCQA questions that the model failed to output the correct answers and realize the following issues:

\textbf{(1) Logical and mathematical reasoning} is a common error that LLMs such as InstructGPT often make \cite{gao2022pal}. Case 1 provides an illustration of this phenomenon, where the model correctly identifies the required operands and operators, but still produces incorrect computational output. The model also frequently makes logical errors, as demonstrated in Case 2, where despite correctly calculating the average percentage of three bars as 6.75\%, which is greater than 6\%, the model fails to conclude the correct answer.

\textbf{(2) Visual data table generation challenge} Although directly fine-tuning UniChart \cite{masry2023unichart} to generate visual data tables with correct positions gives us some minor improvements, we observe that UniChart sometimes generates wrong names of the colors. \Cref{tab:sample-output-vdtg} illustrates some of such cases. 
Given a huge room for future research as discussed in \Cref{appdx:analysis-visual-information}, future works can focus on improving this module.

\paragraph{$\bullet$}{\textbf{Long-form Chart Question Answering \& Chart Summarization}}
Our human evaluation on {\model} reveal several key challanges:

\textbf{(1) Hallucinations \& factual errors} are observed from the generated answers of our \model. These errors are well-known LLMs issues, and have been studied extensively by prior works \cite{fabbri2022qafacteval,ji2023survey}. For LCQA, one example of such errors is Case 5 in which the model generates unwanted numbers (highlighted in \textcolor{red}{red} color) that are not in the chat image. Similarly for CS, in Case 6, the \textcolor{red}{red} passage could not be found in the chart image and input information, but the model generates it confidently.

\textbf{(2) Logical \& mathematical reasoning errors} also appear during generating long-form answers and summaries. Case 4 is an example of LCQA task, in which "decreased by 10 percentage points" should be from "88 \% to 78 \%" instead of "78 \% to 88 \%" as generated. Additionally, Case 3 illustrates a CS example, in which the model, despite correctly retrieving 8269.54 kg in 2019, a number being much bigger than 6348.43 in 2011, still concludes that 6348.43 in 2011 is the largest amount.




\section{Conclusion}
In this work, we take the first step to study few-shot prompting with large language models for chart question answering and summarization tasks. We propose \model{}, a multimodal few-shot prompting framework with LLMs. Through meticulous task analysis, we have developed a comprehensive set of prompting guidelines that are specifically tailored to each downstream task. These guidelines aim to maximize the few-shot performance of large language models (LLMs). Additionally, we propose the \emph{visual data table} representation that incorporates visual features into the prompt and the LLM. Experimental results show that {\model} achieves significant improvements over prior one-shot and fine-tuned baselines, as evidenced by automatic metrics and human evaluation. In the future, we plan to explore dynamically selecting the demonstrations in the prompt to further optimize the performance of LLMs.

\section*{Limitations} One limitation of our work is the utilization of a fixed prompt for each task, which has the potential to restrict the model's performance. To mitigate this issue, we made our best to explicitly set our proposed properties to cover the different challenges in the LCQA and CS benchmarks. Still, more investigation is needed to explore different prompt setups that might improve performance. 
In the factoid chart question answering task, our analysis of the challenges and incorporating them into our fixed prompt serves as an additional effort to address this issue. In addition, our manual human evaluations in \Cref{sec:error-analysis} reveal that InstructGPT often makes logical \& mathematical reasoning errors, as well as hallucinations \& factual errors during generating long-form texts. In the future, we will focus on improving the VDTG module to make it more robust to different types of charts, as well as improving the logical and mathematical reasoning capabilities of LLMs.


\section*{Ethics Statement}
\subsection{Risk in Model Deployment} Deploying a technology developed by a third party such as OpenAI carries ethical considerations that must be addressed to ensure fairness and mitigate potential harm. This involves choosing algorithms and performance metrics that promote fairness \cite{corbett2017algorithmic}, as well as identifying and addressing any biases in the pre-trained model. In this work, we evaluate the performance of the models by a factual correctness metric \cite{fabbri2022qafacteval}, which encourages the generated texts to be factually correct based on the input information and discourages hallucinations. However, it is important to acknowledge that although our method strives for factual correctness, it does not guarantee perfectly factual outputs and may be abused to spread misinformation among the public.
In addition, through human evaluations, we observe that our proposed method does not generate any discriminatory or insulting responses. However, it is important to note that we cannot guarantee that our method will never produce harmful content. Vigilance is necessary to mitigate the risk of producing harmful outputs. 

\subsection{Human Evaluation} During the human evaluation experiments, we hired three annotators to score a total of 300 generated texts. To fairly compensate them, we paid them an hourly wage of \$15, which is higher than the local statutory minimum wage. Additionally, to protect the privacy of the annotators, we kept all annotations anonymized.


\section*{Acknowledgements}
This research was supported by the Natural Sciences \& Engineering Research Council (NSERC) of Canada and Canada Foundation for Innovation (CFI). Do Xuan Long is supported by the A*STAR Computing and Information Science (ACIS) scholarship, A*STAR, Singapore.

\bibliographystyle{acl_natbib}
\bibliography{chart2text}
\newpage
\input{emnlp2020-templates/Appendix}

\end{document}

%% file: emnlp2020-templates/Appendix.tex
\appendix
\section{Appendices}

\subsection{Extended Related Work} \label{addpx:extended-related-work}

\subsubsection{Chart Data Extraction} 
Information Extraction (IE) from charts is a challenging task due to variations in chart types and styles. Early works in this area use a combination of different approaches to creating a pipeline that extracts data from chart \citep{poco2017reverse,balaji2018chart}. These feature-based approaches are shown to not generalize well on various kinds of charts. Therefore, more recent works adopt deep learning models to extract data from charts e.g. \citet{choi2019visualizing} use the idea of detecting bounding boxes for bar charts. \citet{liu2019data} propose a recurrent network for pie charts. These methods achieved significant improvements. However, they are specifically designed for one type of chart. To overcome this, \citet{luo2021chartocr} proposes a hybrid framework that can generate data from different types of charts. ChartOCR \citep{luo2021chartocr} achieves the SOTA results at the time, but similar to other pipeline-based methods, the error will propagate through the pipeline, hence can affect the performance drastically. As end-to-end approaches in other computer vision tasks gained attention, \citet{liu2022matcha} propose an end-to-end model which was based on Pix2Struct \citep{lee2022pix2struct} architecture and further tuned on chart derendering and math reasoning. Recently, \citet{masry2023unichart} propose \emph{UniChart}, an end-to-end pretrained chart model which achieves state-of-the-art result on chart-to-table generation task. 
As the focus of this paper is on constructing prompts for the chart comprehension tasks, further exploration of end-to-end data extraction approaches is left for future works.

\subsection{VDTG Training Details} \label{appdx:vdtg-training-details}
We fine-tune UniChart \cite{masry2023unichart} model for 50K steps with a batch size of 80. We use a learning rate of 1e-4 and 500 warmup steps. We evaluate the model every 1K steps and choose the best checkpoint for inference.

\subsection{FCQA Prompt} \label{appdix:chartqa-prompt}

\Cref{tab:chartqa-demonstrations} illustrates our full demonstrations for the task of factoid chart question answering. The samples are selected carefully from the validation split of ChartQA \cite{masry-etal-2022-chartqa}, each of them covers one question type that the InstructGPT model fails to answer presented in \Cref{fiq:ablation-prompt-chartqa}.

\begin{table*}[t!]
  \centering
  \small
  \scalebox{.9}{
    \begin{tabular}{p{1cm}|p{4cm}|p{8cm}|p{2.5cm}}
    \toprule
    Index & Question & CCR Format: \texttt{\textcolor{olive}{operands with visual attributes}, \textcolor{magenta}{operands without visual attributes}, \textcolor{blue}{operators}, \textcolor{ao(english)}{reasoning}, \textcolor{violet}{result}} & Question Type \\
    \midrule
    1 & How many data points on the disapprove line are above 50? & \textcolor{magenta}{The disapprove number 51 and 50} \textcolor{ao(english)}{are greater than 50}. \textcolor{violet}{The answer is 2}. & Complex Retrieval \\
    2 & How much value of 'Mostly bad news' increased from Jan. 2011 to Sept. 2011? & \textcolor{magenta}{The value of 'Mostly bad news' in Jan. 2011 is 24 and in Sept. 2011 is 61}. So \textcolor{blue}{the increase is 61 - 24 = 37}. \textcolor{violet}{The answer is 37}. & Add/Substract \\
    3 & What's the ratio of Lean Republican segment and Republican segment? & \textcolor{magenta}{The value of Lean Republican segment is 39 and the value of Republican segment is 53}. \textcolor{blue}{The ratio is 39 / 53 = 0.7358}. \textcolor{violet}{The answer is 0.7358}. & Divide \\
    4 & What's the average value of the first two blue bars in the chart? & \textcolor{olive}{The first two blue bars are 30-34 years old with 0.35 and 25-29 years old with 0.31}. \textcolor{blue}{The average value is ( 0.35 + 0.31 ) / 2 = 0.33}. \textcolor{violet}{The answer is 0.33}. & Visual Reasoning \\
    5 & What's the total sum of peak points of all three lines? & \textcolor{olive}{The peak point of red line is 80, the peak point of yellow line is 68, the peak point of green line is 7}. \textcolor{blue}{The total sum of peak points is 80 + 68 + 7 = 155}. \textcolor{violet}{The answer is 155}. & Compositional Reasoning \\
    6 & Is the average value of Andean Latin America and Cambodia more than the value of Thailand? & \textcolor{magenta}{The value of Andean Latin America is 1.47 and the value of Cambodia is 0.77}. So \textcolor{blue}{the average value of Andean Latin America and Cambodia is ( 1.47 + 0.77 ) / 2 = 1.12}. \textcolor{magenta}{The value of Thailand is 0.39}. Since \textcolor{ao(english)}{1.12 > 0.39, the average value of Andean Latin America and Cambodia is more than the value of Thailand}. \textcolor{violet}{The answer is Yes}. & Boolean \\
    \bottomrule
    \end{tabular}}
  \vspace{-2mm}
  \caption{\small{Demonstrations for factoid chart question answering task. Samples are from the validation split of ChartQA \cite{masry-etal-2022-chartqa} dataset.}} 
  \label{tab:chartqa-demonstrations}
\end{table*}


\subsection{LCQA Prompt} \label{appdix:opencqa-prompt}
We present all the demonstrations used for LCQA task in \Cref{tab:opencqa-demonstrations}. The 6 samples are selected carefully from the validation split of OpenCQA \cite{opencqa} dataset.

\begin{table*}[t!]
  \centering
  \small
  \scalebox{.73}{
    \begin{tabular}{p{1cm}|p{10cm}|p{7cm}|p{2cm}}
    \toprule
    Index & Input & Output & Question Type \\
    \midrule
    1 & What opinion do the public have about earth warming? Title: Is there solid evidence the earth is warming? Mixed evidencel Total U.S population Context: Is there solid evidence the earth is warming?  <s> Mixed evidencel  <s> Yes. because of  <s> Yes. because of  <s> Yes but don't  <s> No  <s> Some evidencel  <s> human activity  <s> natura patterns  <s> know cause  <s> Don't know  <s> Total U.S population  <s> 47 \%  <s> 18\%  <s> 6\% 21 \%  <s> 8\%  <s> Unaffliated  <s> 58\%  <s> 11\%  <s> 6\% 18\%  <s> 7\%  <s> White mainline Protestants  <s> 48\%  <s> 19\%  <s> 6\% 19\%  <s> 8\%  <s> White non-Hispanic Catholics  <s> 44\%  <s> 20\%  <s> 6\% 22\%  <s> 8\%  <s> Black Protestants  <s> 39\%  <s> 36\%  <s> 5\% 15\%  <s> 5\%  <s> White evangelical Protestants  <s> 17\%  <s> 7\%  <s> 11\%  <s> 34\%  <s> 31\%  <s> The Pew Forum on Religion \& Public Life  <s> pewforum. ORG  
    & 
    The unaffiliated ( 58 \% ) are the most likely among the religious groups studied to say there is solid evidence the earth is warming because of human activity . White evangelical Protestants are the most likely to say there is no solid evidence the earth is warming ( 31 \% ) , and the least likely to believe that humans have contributed to heating up the planet ( 34 \% ) . While only 39 \% of black Protestants say global warming is a result of human activity , they are , however , the least likely of the religions studied to deny global warming is occurring ( 15 \% ) .
    & Describe and Summary \\
    \midrule
    2 & Among families with opt-out moms, how is the difference in education between husbands and wives? Title: About One-in-Ten Highly Educated Mothers 'Opt Out' Among all educated moms 'opt-out' moms Context: About One-in-Ten Highly Educated Mothers "Opt Out"  <s> Among all highly  <s> Among all  <s> educated moms  <s> "opt-out" moms  <s> 9\%  <s> are "opt-out"  <s> moms  <s> 37\% More education  <s> than husband  <s> 45  <s> Same  <s> education  <s> as hus band  <s> 18  <s> Less education  <s> than husband  <s> "Highly educated mothers have at least a Master's de gree, are ages 18 69 and  <s> are living with the ir own child(ren) working younger than 18. family Among these moms, "opt-out"  <s> mothers living are married with working hus band, annual family Among income above \$7 5, .000  <s> and say they are not working in forder to care for family. a professional degree is  <s> considered equivalent to a Ph.D for the purpose of the compa rison of spousal  <s> educational attainment.  <s> Source: Pew Research Center analysis of March Current Population Surveys  <s> Integrated Public Use Microdata Series (IPUMS CPS), 2012 2013  <s> PEW RESEARCH CENTER 
    & In 37 \% of the cases , it is the stay - at - home wives who actually have a higher level of education . In 45 \% of these families , the spouses have equal educational attainment , and in about 18 \% of the cases , the husbands have more education than their wives .
    & Comparative \\
    \midrule
    3 & What was the trend among registered Hispanic voters about Trump's work? Title: Most Hispanic voters disapprove of Trump, dissatisfied with nation's direction Context: Most Hispanic voters disapprove of Trump, dissatisfied with nation's direction  <s> \% who are country with the today way things are  <s> \% who job president of the way Trump is handling  <s> going in the country today  <s> his job as president  <s> Disapprove  <s> Approve  <s> Not so  <s> Not so  <s> Very strongly strongly Very  <s> Dissatisfied  <s> Satisfied  <s> NET  <s> NET  <s> strongly  <s> strongly  <s> All Hispanic RVs  <s> 68  <s> 51  <s> 17 8 23  <s> 30  <s> 67  <s> 32  <s> Dem/Lean Dem  <s> 93  <s> 73  <s> 20 3 4  <s> 80  <s> 19  <s> 1211 17  <s> 75  <s> Rep/Lean Rep  <s> 22  <s> 58  <s> 45  <s> 54  <s> Note: Based on Hispanic registered voters. Share of respondents who didn't offer an answer not shown. Figures may not add to NET due to  <s> rounding.  <s> Source: National Survey of Latinos conducted Dec. 3-23, 2019  <s> PEW RESEARCH CENTER  
    & About two - thirds of Hispanic registered voters ( 68 \% ) disapprove of the job Trump is doing as president , including 51 \% who disapprove very strongly . The 30 \% of Hispanic voters who approve of Trump includes 23 \% who approve strongly & Discover \\
    \midrule
    4 & Describe the trend in job availability to people over the past years. Title: Job Availability in Your Area 2001 2008 Context: Job Availability in Your Area  <s> 2001 2008  <s> Difficult to find  <s> Plenty of jobs  <s> 66  <s> 60  <s> 480  <s> 40  <s> 20  <s> Q  <s> 2001 2002 2003 2004 2005 2006 2007 2008 
    & Overall , a majority ( 53 \% ) says that jobs are difficult to find in their community while only about a third ( 34 \% ) says there are plenty of jobs available . The percentage saying that jobs are difficult to find locally has increased modestly since February 2007 ( 48 \% ) . In October 2003 , 66 \% said that jobs were difficult to find , the highest percentage expressing that view since the start of the Bush administration .
    & Describe and Summary \\
    \midrule
    5 & How was the public's attention on different news? Title: News Interest vs. News Coverage Context: News Interest vs. News Coverage  <s> April 13-19  <s> Pirates  <s> 16  <s> 34  <s> Economy 27  <s> 18  <s> C  <s> Tea parties  <s> 3  <s> 6  <s> Obama abroad  <s> 3  <s> Cuba policy  <s> 2  <s> CIA memos  <s> Interest: percent who named story as most closely followed  <s> Coverage: percent of news coverage devoted to story  
    & 34 \% say they followed stories about the continued attempts by pirates to hijack ships more closely than any other story last week , while 27 \% say they followed stories about the U.S. economy most closely . 9 \% say they followed the tea party protests more closely than any other story .
    & Comparative \\
    \midrule
    6 & Does China believe that Japan has sufficiently apologized for their actions during WW2? Title: Chinese unchanged in belief that Japan has yet to sufficiently apologize for actions during World War I \% Chinese saying Japan has for its military actions during the 1930s and 1940S Context: Chinese unchanged in belief that Japan has yet to  <s> sufficiently apologize for actions during World War I  <s> \% Chinese saying Japan has for its military actions during the 1930s and  <s> 1940S  <s> Apologized No apology  <s> Not apologized  <s> sufficiently  <s> sufficiently necessary  <s> 2016  <s> 77\%  <s> 10\%  <s> 2\%  <s> 2013  <s> 78  <s> 2008  <s> 76  <s> 8  <s> 2006  <s> 81  <s> Source: Spring 2016 Global Attitudes Survey Q84.  <s> PEW RESEARCH CENTER 
    & Roughly three - quarters ( 77 \% ) say Japan has not adequately expressed regret , and such Chinese sentiment is largely unchanged since 2006 . Only 10 \% believe Tokyo has apologized enough . & Discover \\
    \bottomrule
    \end{tabular}}
  \vspace{-2mm}
  \caption{\small{Demonstrations for long-form chart question answering task. Samples are from the validation split of the OpenCQA \cite{opencqa} dataset. The chart images are provided in \Cref{tab:opencqa-demonstration-consistency}.}}
  \label{tab:opencqa-demonstrations}
\end{table*}

\subsection{Chart Summarization Prompt} \label{appdix:chart-to-text-prompt}
\Cref{tab:chart2text-demonstrations} presents all the demonstrations we use for the task of chart summarization. All 6 demonstrations are from Chart-to-Text dataset \cite{kantharaj-etal-2022-chart}. The first three samples are taken from validation set of Pew dataset, whilst the last 3 from Statista dataset.

\begin{table*}[t!]
  \centering
  \small
  \scalebox{.73}{
    \begin{tabular}{p{1cm}|p{9cm}|p{8cm}|p{2cm}}
    \toprule
    Index & Input & Output & Summarization Type \\
    \midrule
    1 & Seven-in-ten Americans said in 2019 that their personal info was less secure than five years earlier \% ofU.S adults who said they felt as iftheir personal information was than it was five years ago <SEP> Seven-in-ten Americans said in 2019 that their personal info was less secure than five years earlier \% ofU.S adults who said they felt as iftheir personal information was than it was five years ago More About Less secure the same secure 6 24 70 Note: Those who did not give an answer are not shown. Source: Survey conducted June 3- 17, 2019. "Americans and Privacy: Concerned, Confused and Feeling Lack of Control Over Their Personal Information PEW RESEARCH CENTER 
    & Before the outbreak, Americans strongly believed their personal data was more vulnerable than in the past. In a June 2019 survey, 70\% of Americans said their personal information was less secure than it was five years earlier. Just 6\% of Americans said they felt their information was more secure than in the past, while 24\% said their personal information was about as secure as it was five years earlier. Experts have suggested that data security could be threatened for companies shifting to remote work – as well as ordinary people who are spending more time online – during the coronavirus outbreak.
    & Perceptual and Cognitive \\
    \midrule
    2 & Figure by Race, Ethnicity and Citizenship, 2007 <SEP> Figure  Composition of Sentenced Federal Offenders,  by Race, Ethnicity and Citizenship, 2007  196 ofsentenced offenders)  U.S citizen  Non-U.S citizen  40  29  25  21  Hispanic  White  Black  Notes: Percenta ges may not total due to rounding. The category  "Other" is excluded  Source: Pew Hispanic Center analysis of United States Sentencing  Source: Commission Monitoring offederal Criminal Sentences data, 2007 
    & Sharp growth in illegal immigration and increased enforcement of immigration laws have dramatically altered the ethnic composition of offenders sentenced in federal courts. In 2007, Latinos accounted for 40\% of all sentenced federal offenders‚Äîmore than triple their share (13\%) of the total U.S. adult population. The share of all sentenced offenders who were Latino in 2007 was up from 24\% in 1991, according to an analysis of data from the United States Sentencing Commission (USSC) by the Pew Hispanic Center, a project of the Pew Research Center.
    & Visual Encodding \\
    \midrule
    3 & Half of voters, including a majority of Biden supporters, expect it will be difficult to vote in this year's election \% ofregistered voters wwo expect voting in the November elections to be <SEP> Half of voters, including a majority of Biden supporters,  expect it will be difficult to vote in this year's election  \% ofregistered voters wwo expect voting in the November elections to be  Republican Support  Democratic Support  All voters  Republican Support  Democratic Support  candidate  candidate  90  Easy  85  83  64  50  40  10  Difficult  15  17  35  49  60  Oct  Aug  Oct  Aug  Oct  Aug  18  -20  '18  20  1'8  --20  House  Presidential  election  election  Notes: Based on registered voters Based on supporters of Republican fand Democratic  House candidates lin 2018 Trump and Biden supporters in 2020 No answer responses snot  shown.  Source: Survey of U.S adults conducted July 27-Aug 2 2020  PEW RESEARCH CENTER
    & Far more voters expect voting to be difficult this year than in 2018, according to a survey of registered voters conducted July 27 to Aug. 2. Around half of registered voters (49\%) said they expect voting to be very or somewhat difficult this year, while the other half (50\%) expect it to be very or somewhat easy. That represents a major change from October 2018, when just 15\% of voters said they expected voting in that year‚Äôs midterms to be difficult and 85\% said they expected it to be easy.
    &  Statistical and Comparative \\
    \midrule
    \midrule
    4 & Turnover from retail sale of bakery products and cereals ( including rice and pasta ) in the United Kingdom ( UK ) from 2008 to 2018 ( in million GBP ) Context: Year \& Turnover in million GBP <SEP> 2018 \& 21890 <SEP> 2017 \& 21433 <SEP> 2016 \& 20757 <SEP> 2015 \& 22187 <SEP> 2014 \& 20372 <SEP> 2013 \& 20352 <SEP> 2012 \& 20078 <SEP> 2011 \& 18818 <SEP> 2010 \& 17875 <SEP> 2009 \& 16719 <SEP> 2008 \& 15816 . 
    & This statistic shows the total annual turnover of the retail sale of bakery products and cereals ( including rice and pasta products ) in the United Kingdom from 2008 to 2018 . Turnover has grown steadily , reaching a peak of approximately 22 billion British pounds in 2015 before declining the following year . By 2018 turnover had almost completely recovered and amounted to 21.9 billion pounds . 
    & Perceptual and Cognitive  \\
    \midrule
    5 & Number of nights spent camping or in caravans in Great Britain in 2019 , by type ( in millions ) Context: type \& Nights spent in millions <SEP> Glamping/alternative accommodation e.g. yurt, tipi, ecopod \& 1.7 <SEP> Static caravan (owned) \& 9.4 <SEP> Tent \& 12.5 <SEP> Static caravan (rented) \& 16.8 <SEP> Touring caravans (towed/campervan/motorhome) \& 19.8
    & This statistic shows the number of nights spent camping or in caravans in Great Britain in 2019 , by type . Almost 20 million nights were spent by domestic tourists in touring caravans in Great Britain , while 12.5 million nights were spent camping in tents . & Statistical and Comparative \\
    \midrule
    6 & Barbados : National debt in relation to gross domestic product ( GDP ) from 2015 to 2025 Context: Year \& National debt in relation to GDP <SEP> 2025* \& 100.11\% <SEP> 2024* \& 105.68\% <SEP> 2023* \& 111.34\% <SEP> 2022* \& 117.24\% <SEP> 2021* \& 124.53\% <SEP> 2020 \& 134.09\% <SEP> 2019 \& 122.22\% <SEP> 2018 \& 125.59\% <SEP> 2017 \& 158.26\% <SEP> 2016 \& 149.45\% <SEP> 2015 \& 147.02\% 
    & 
    This statistic shows the national debt of Barbados from 2015 to 2020 in relation to the gross domestic product ( GDP ) , with projections up until 2025 . The figures refer to the whole country and include the debts of the state , the communities , the municipalities and the social insurances . In 2020 , the national debt of Barbados amounted to approximately 134.09 percent of the GDP .
    & Visual Encodding \\
    \bottomrule
    \end{tabular}}
  \vspace{-2mm}
  \caption{\small{Demonstrations for chart summarization task. Samples are obtained from the validation split of Chart-to-Text \cite{kantharaj-etal-2022-chart} dataset with the first three from Pew, and the last three from Statista. All the chart images are provided in \Cref{tab:chart2text-demonstrations-consistency}.}}
  \label{tab:chart2text-demonstrations}
\end{table*}


\subsection{Visual Data Table Generator (VDTG)} \label{subsubsec:visual-data-table-generator}
\begin{figure*}[t!]
     \centering
        \includegraphics[width=1\linewidth, trim={0cm 0cm 0cm 0cm},clip]{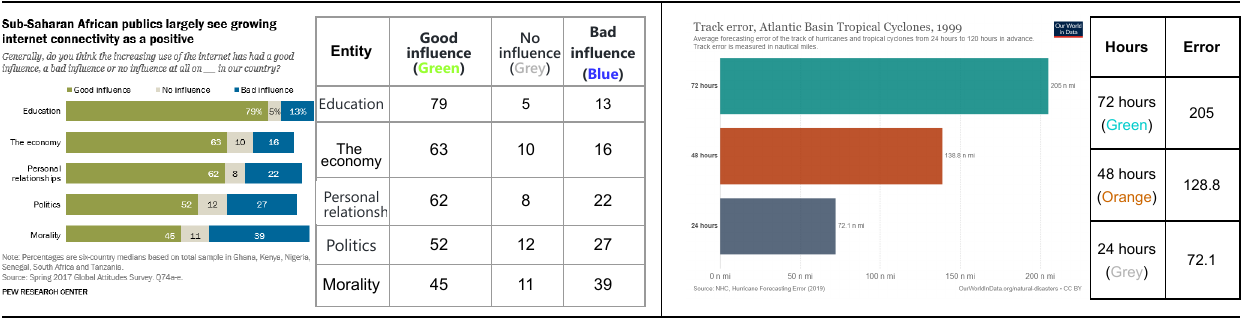}
         \caption{\small{Two types of visual data tables. The left-hand side is \emph{horizontal} whilst the right-hand one is \emph{vertical}. 
         } 
         }
    \label{fiq:visual-data-table}
\end{figure*}

There are two types of visual data tables that we constructed to train the model, as shown in \Cref{fiq:visual-data-table}.

\Cref{tab:sample-output-vdtg} shows some samples outputs from our proposed VDTG module. We observe that the first two cases the VDTG model can generate the correct colors and positions of the labels. In the $3$-rd case, however, the model makes an error when outputting the "light blue" instead of "blue". In the $4$-th case, the model outputs the correct colors but wrong positions since the line "Women" should come first. In the $5$-th case, it is challenging for the model to distinguish between the colors that are similar. In the $6$-th case, the model also makes an error in generating blue instead of the correct color is gray.

\begin{table*}[t!]
  \centering
  \small
  \scalebox{.65}{
    \begin{tabular}{p{1cm}|p{7cm}|p{10cm}}
    \toprule
    Index & Chart Image & Generated Visual Data Table \\
    \midrule
    1 & 
    \raisebox{-1\height}{\hspace{0mm}  \includegraphics[width=.4\textwidth,trim={0cm 0cm 0cm 0cm},clip]{emnlp2020-templates/case-studies/58486764007015.png}}
    & Country | Human Rights Volations, 2012 <0x0A> Central African Republic (\textcolor{ao(english)}{red}) | 8.5 <0x0A> Iraq (\textcolor{ao(english)}{purple}) | 8.3 <0x0A> Gabon (\textcolor{ao(english)}{Teal}) | 6.8 <0x0A> Suriname (\textcolor{ao(english)}{orange}) | 5.3 <0x0A> Peru (\textcolor{ao(english)}{gray}) | 4.9 \\
    \midrule
    2 & 
    \raisebox{-1\height}{\hspace{0mm}  \includegraphics[width=.4\textwidth,trim={0cm 0cm 0cm 0cm},clip]{emnlp2020-templates/case-studies/multi_col_20010.png}}
     & Characteristic | Nominal (\textcolor{ao(english)}{blue}) | Real Terms (\textcolor{ao(english)}{navy blue}) <0x0A> 2019/20 | 11.97 | 12.11 <0x0A> 2018/19 | 11.35 | 11.57 <0x0A> 2017/18 | 11.94 | 11.76 <0x0A> 2016/17 | 11.6 | 12.25 <0x0A> 2015/16 | 11.34 | 12.3 <0x0A> 2014/15 | 12.46 | 13.55 <0x0A> 2013/14 | 11.62 | 12.97 <0x0A> 2012/13 | 12.76 | 14.43 <0x0A> 2011/12 | 12.57 | 14.35 <0x0A> 2010/11 | 13.06 | 15.24 <0x0A> 2009/10 | 13.22 | 15.76 <0x0A> 2008/09 | 12.46 | 15.03 <0x0A> 2007/08 | 11.94 | 14.76 <0x0A> 2006/07 | 11.43 | 14.48 <0x0A> 2005/06 | 10.87 | 14.21 <0x0A> 2004/05 | 10.03 | 13.36 <0x0A> 2003/04 | 9.76 | 13.3 <0x0A> 2002/03 | 9.37 | 13.03 <0x0A> 2001/02 | 8.66 | 12.42 <0x0A> 2000/01 | 7.89 | 11.38 <0x0A> 1999/00 | 7.74 | 11.48 <0x0A> 1998/99 | 7.22 | 10.86 <0x0A> 1997/98 | 6.41 | 9.76 <0x0A> 1996/97 | 5.73 | 8.8 \\
     \midrule
    \midrule
    3 & 
    \raisebox{-1\height}{\hspace{0mm}  \includegraphics[width=.4\textwidth,trim={0cm 0cm 0cm 0cm},clip]{emnlp2020-templates/case-studies/3960.png}}
     & Frequently (\textcolor{ao(english)}{dark blue}) | Sometimes (\textcolor{red}{light blue}) | NET <0x0A> Lonely | 7 | 24 | 31 <0x0A> Depressed | 13 | 36 | 49 <0x0A> Inspired | 16 | 53 | 69 <0x0A> Connected | 21 | 49 | 71 <0x0A> Angry | 25 | 47 | 71 <0x0A> Amused | 44 | 44 | 88\\
     \midrule
    4 & 
    \raisebox{-1\height}{\hspace{0mm}  \includegraphics[width=.4\textwidth,trim={0cm 0cm 0cm 0cm},clip]{emnlp2020-templates/case-studies/multi_col_80264.png}}
     & Characteristic | Men (\textcolor{ao(english)}{blue}) | Women (\textcolor{ao(english)}{navy blue}) <0x0A> 2019 | 945.74 | 120.01 <0x0A> 2018 | 928.34 | 1181.71 <0x0A> 2017 | 906.45 | 1160.94 <0x0A> 2016 | 894.49 | 1154.19 <0x0A> 2015 | 895.74 | 1150.61 <0x0A> 2014 | 892.76 | 1156.91 <0x0A> 2013 | 880.11 | 1134.59 <0x0A> 2012 | 863.35 | 1118.29 <0x0A> 2011 | 846.25 | 1001.84 <0x0A> 2010 | 818.71 | 1069.38 <0x0A> 2009 | 766.65 | 1021.74 <0x0A> 2008 | 751.52 | 998.79 <0x0A> 2007 | 733.65 | 983.84 <0x0A> 2006 | 725.31 | 971.54 <0x0A> 2005 | 719.52 | 959.92 <0x0A> 2004 | 708.46 | 941.77 <0x0A> 2003 | 669.77 | 887.1) <0x0A> 2002 | 645.65 | 852.11 <0x0A> 2001 | 619.74 | 817.46 <0x0A> 2000 | 587.16 | 756.59 \\
    \midrule
    5  & 
    \raisebox{-1\height}{\hspace{0mm}  \includegraphics[width=.4\textwidth,trim={0cm 0cm 0cm 0cm},clip]{emnlp2020-templates/case-studies/5967.png}}
     & Entity | More (\textcolor{red}{green}) | Aboutthe same (\textcolor{red}{gray}) | Fewer (\textcolor{red}{orange}) | None (VOL) <0x0A> Turkey | nan | 30.0 | 53.0 | 0 <0x0A> Christian | 10.0 | 33.0 | 53.0 | 0 <0x0A> Sunni | 12.0 | 46.0 | 36.0 | 0 <0x0A> Shia | nan | nan | 44.0 | 33 <0x0A> Lebanon | 13.0 | 40.0 | nan | 0 <0x0A> Jordan | nan | nan | 63.0 | 27 <0x0A> Syrian | 64.0 | nan | nan | 26 <0x0A> Jordan | 8.0 | 8.0 | 60.0 | 23 \\
    \midrule
    6 & 
    \raisebox{-1\height}{\hspace{0mm}  \includegraphics[width=.4\textwidth,trim={0cm 0cm 0cm 0cm},clip]{emnlp2020-templates/case-studies/26027711018261.png}}
     & Country | Scheduled teaching time that teachers are in the classroom, 2002 to 2016 <0x0A> Egypt (\textcolor{ao(english)}{gray}) | 93.45 <0x0A> Tunisia (\textcolor{red}{blue}) | 89.89 <0x0A> Madagascar (\textcolor{red}{blue}) | 58.09 <0x0A> Mozambique (\textcolor{red}{blue}) | 44.0 \\
    \bottomrule
    \end{tabular}}
  \vspace{-2mm}
  \caption{\small{Sample outputs of our VDTG module. The first two cases the model detects the correct colors and positions, while the last 4 cases the model fails.}}
  \label{tab:sample-output-vdtg}
\end{table*}

\subsection{Consistency of Generated Output with Demonstrations} \label{sec:consistency-generated-demonstration}
There have one consistency property that we propose in our prompt construction rules for LCQA and CS tasks (\Cref{sec:method}): \emph{(iii)} in \Cref{sec:method}. Although generating texts given a piece of information is a one-to-many relationship, i.e., there may have multiple correct answers, our motivation for this property is that the language model should answer consistently and correctly what it is taught through demonstrations. To verify this property in our proposed prompts, for each of LCQA and CS tasks, we input the proposed prompt and ask the model to generate the answer for each question in the prompt. The experimental results for LCQA and CS are illustrated in \Cref{tab:opencqa-demonstration-consistency} and \Cref{tab:chart2text-demonstrations-consistency} respectively. We observe that although the textual outputs of the model sometimes are slightly different from the input demonstrations, the main points are kept without adding any hallucinating information, which verifies that our prompts satisfy \emph{(iii)} (\Cref{sec:method}) property.

\begin{table*}[t!]
  \centering
  \small
  \scalebox{.6}{
    \begin{tabular}{p{7cm}|p{7cm}|p{7cm}}
    \toprule
    Chart Image & Demonstration's Output & Generated Output \\
    \midrule
    \raisebox{-1\height}{\hspace{0mm}  \includegraphics[width=.4\textwidth,trim={0cm 0cm 0cm 0cm},clip]{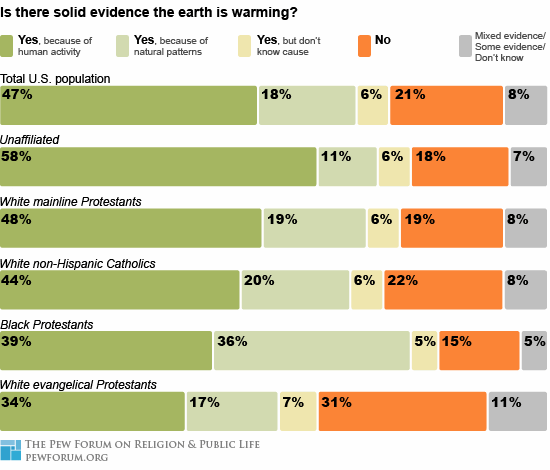}} &  
    The unaffiliated ( 58 \% ) are the most likely among the religious groups studied to say there is solid evidence the earth is warming because of human activity . White evangelical Protestants are the most likely to say there is no solid evidence the earth is warming ( 31 \% ) , and the least likely to believe that humans have contributed to heating up the planet ( 34 \% ) . While only 39 \% of black Protestants say global warming is a result of human activity , they are , however , the least likely of the religions studied to deny global warming is occurring ( 15 \% ) .
    &
    The opinion of the public on earth warming varies among religious groups. The unaffiliated are the most likely among the religious groups studied to say there is solid evidence the earth is warming because of human activity (58\%). White evangelical Protestants are the most likely to say there is no solid evidence the earth is warming (31\%) and the least likely to believe that humans have contributed to heating up the planet (34\%). While only 39\% of black Protestants say global warming is a result of human activity, they are, however, the least likely of the religions studied to deny global warming is occurring (15\%).\\
    \midrule
    
    \raisebox{-1\height}{\hspace{0mm}  \includegraphics[width=.4\textwidth,trim={0cm 0cm 0cm 0cm},clip]{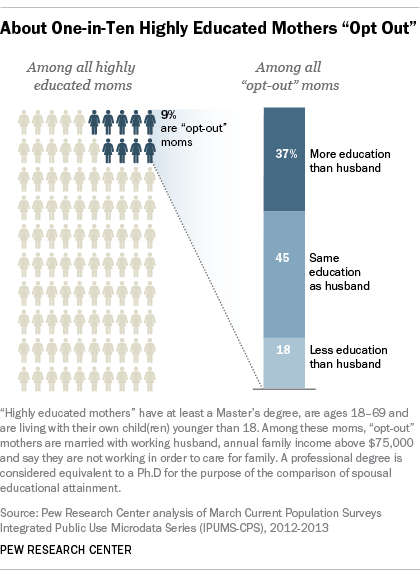}}
    & In 37 \% of the cases , it is the stay - at - home wives who actually have a higher level of education . In 45 \% of these families , the spouses have equal educational attainment , and in about 18 \% of the cases , the husbands have more education than their wives .
    &
    Among families with opt-out moms, 37\% have the stay-at-home wives with a higher level of education than the husbands. 45\% of these families have equal educational attainment between spouses, and 18\% of the cases have the husbands with more education than the wives.\\
    \midrule
    \raisebox{-1\height}{\hspace{0mm}  \includegraphics[width=.4\textwidth,trim={0cm 0cm 0cm 0cm},clip]{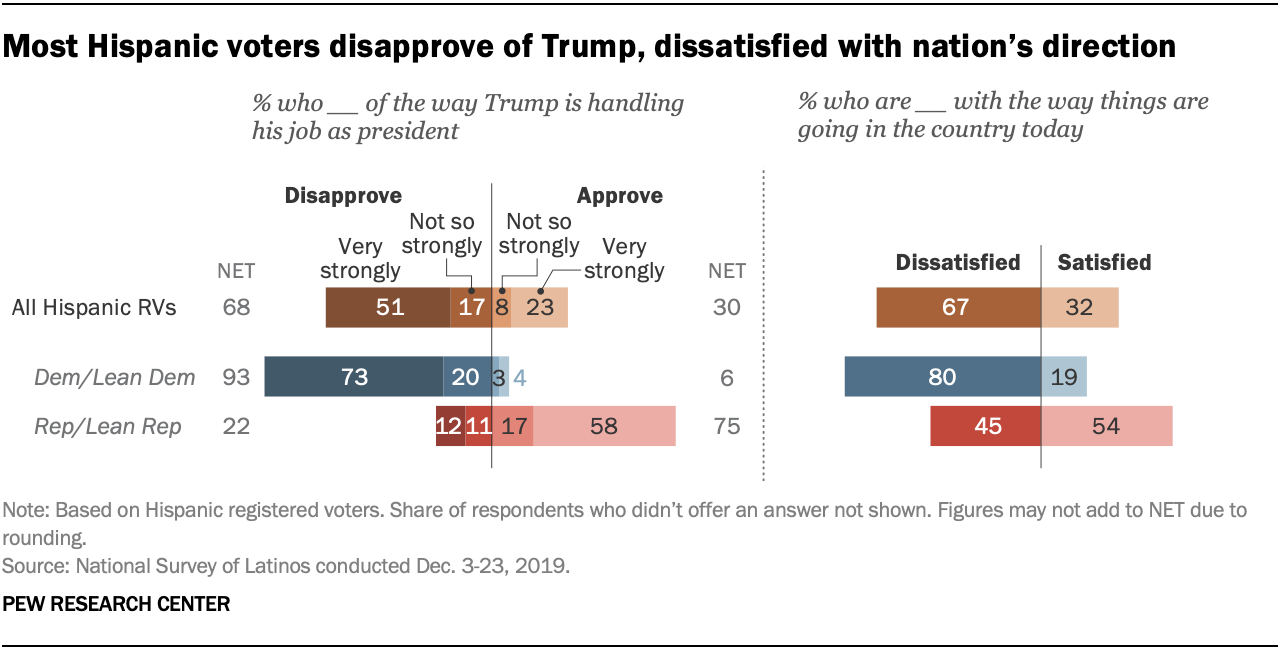}}
    & About two - thirds of Hispanic registered voters ( 68 \% ) disapprove of the job Trump is doing as president , including 51 \% who disapprove very strongly . The 30 \% of Hispanic voters who approve of Trump includes 23 \% who approve strongly 
    &
    Among Hispanic registered voters, there is strong disapproval of President Trump's job performance. About two-thirds (68\%) of Hispanic voters disapprove of the job Trump is doing as president, including 51\% who disapprove very strongly. The 30\% of Hispanic voters who approve of Trump includes 23\% who approve strongly.
    \\
    \midrule
    \raisebox{-1\height}{\hspace{0mm}  \includegraphics[width=.4\textwidth,trim={0cm 0cm 0cm 0cm},clip]{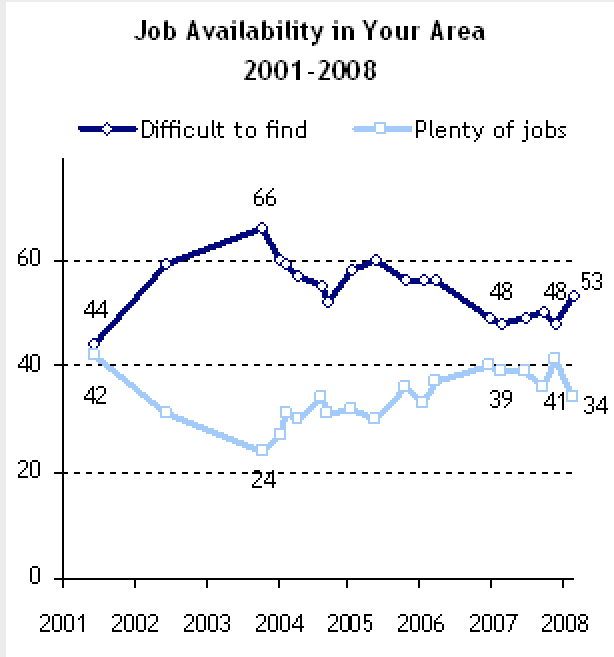}} 
    & Overall , a majority ( 53 \% ) says that jobs are difficult to find in their community while only about a third ( 34 \% ) says there are plenty of jobs available . The percentage saying that jobs are difficult to find locally has increased modestly since February 2007 ( 48 \% ) . In October 2003 , 66 \% said that jobs were difficult to find , the highest percentage expressing that view since the start of the Bush administration .
    &
    Overall, a majority (53\%) says that jobs are difficult to find in their community while only about a third (34\%) says there are plenty of jobs available. The percentage saying that jobs are difficult to find locally has increased modestly since February 2007 (48\%). In October 2003, 66\% said that jobs were difficult to find, the highest percentage expressing that view since the start of the Bush administration.
    \\
    \midrule
    \raisebox{-1\height}{\hspace{0mm}  \includegraphics[width=.4\textwidth,trim={0cm 0cm 0cm 0cm},clip]{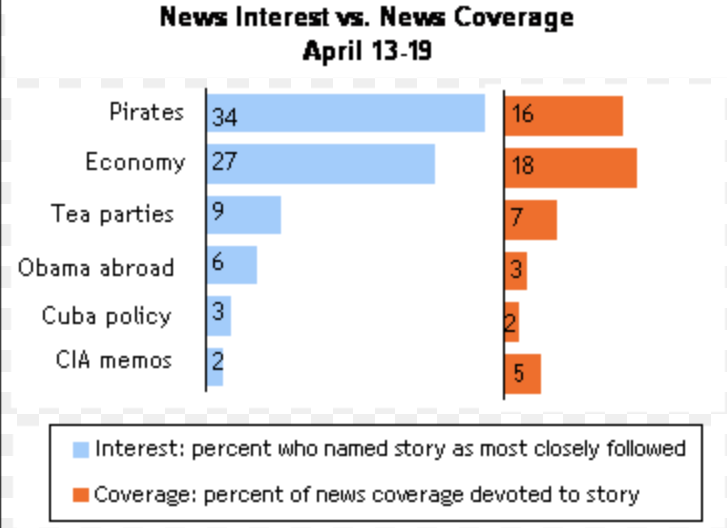}}
    & 34 \% say they followed stories about the continued attempts by pirates to hijack ships more closely than any other story last week , while 27 \% say they followed stories about the U.S. economy most closely . 9 \% say they followed the tea party protests more closely than any other story .
    &
    34\% of the public say they followed stories about the continued attempts by pirates to hijack ships more closely than any other story last week, while 27\% say they followed stories about the U.S. economy most closely. 9\% say they followed the tea party protests more closely than any other story.
    \\
    \midrule
    \raisebox{-1\height}{\hspace{0mm}  \includegraphics[width=.4\textwidth,trim={0cm 0cm 0cm 0cm},clip]{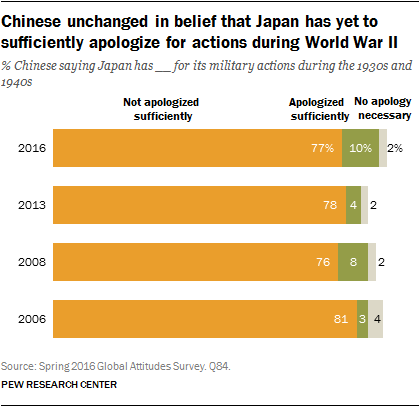}}
    & Roughly three - quarters ( 77 \% ) say Japan has not adequately expressed regret , and such Chinese sentiment is largely unchanged since 2006 . Only 10 \% believe Tokyo has apologized enough . 
    &
    Roughly three-quarters (77\%) of Chinese people say that Japan has not adequately expressed regret for their actions during World War II, and this sentiment has been largely unchanged since 2006. Only 10\% of Chinese people believe that Japan has apologized enough for their actions during the war.
    \\
    \bottomrule
    \end{tabular}}
  \vspace{-2mm}
  \caption{\small{For each demonstration in our prompt, we ask the language model to generate its expected output, given its input. We then compare the generated output from the model with its expected output.
  }}
  \label{tab:opencqa-demonstration-consistency}
\end{table*}

\begin{table*}[t!]
  \centering
  \small
  \scalebox{.6}{
    \begin{tabular}{p{7cm}|p{7cm}|p{7cm}}
    \toprule
    Chart Image & Demonstration's Output & Generated Output \\
    \midrule
    \raisebox{-1\height}{\hspace{0mm}  \includegraphics[width=.4\textwidth,trim={0cm 0cm 0cm 0cm},clip]{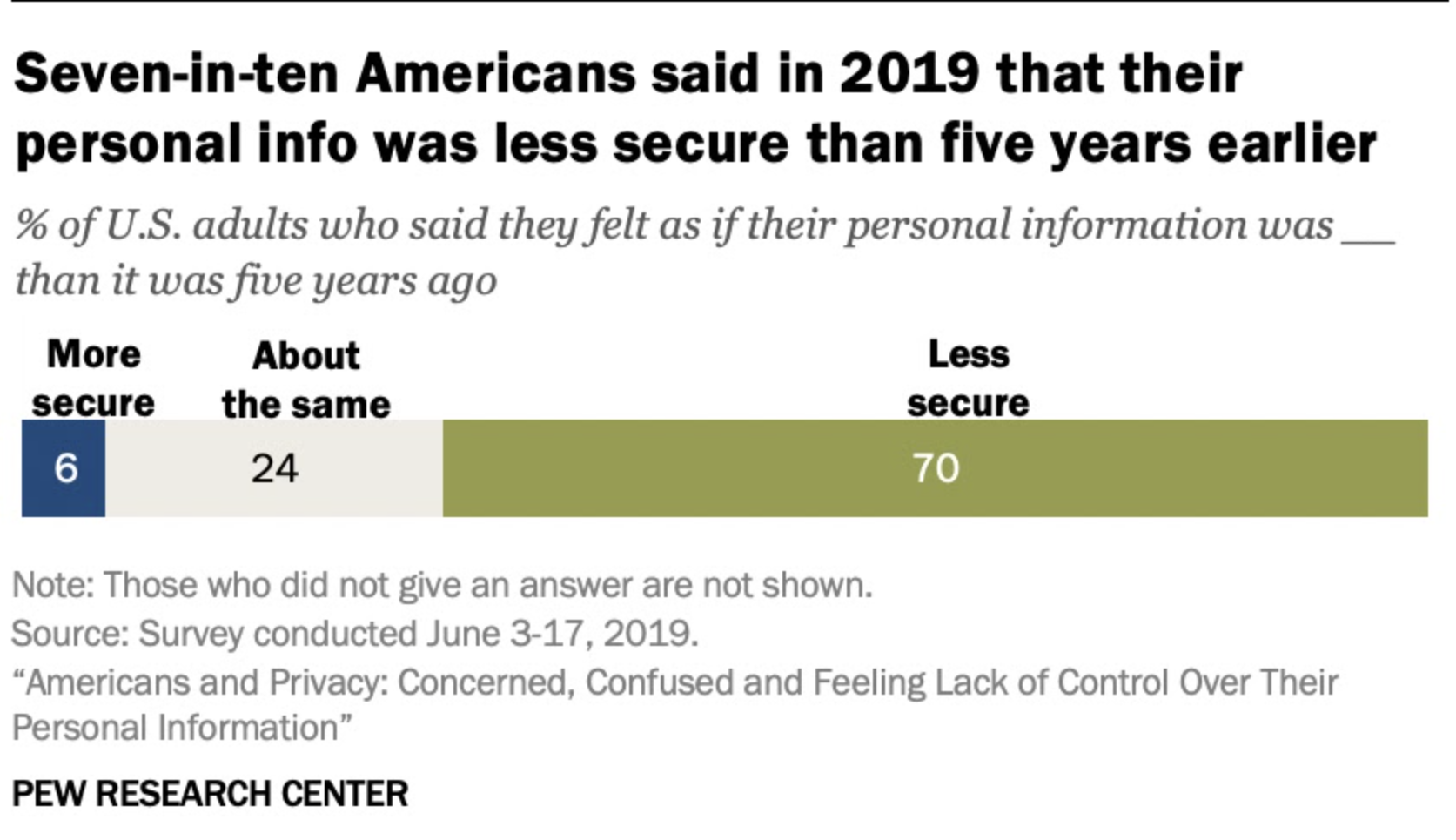}}  
    & Before the outbreak, Americans strongly believed their personal data was more vulnerable than in the past. In a June 2019 survey, 70\% of Americans said their personal information was less secure than it was five years earlier. Just 6\% of Americans said they felt their information was more secure than in the past, while 24\% said their personal information was about as secure as it was five years earlier. Experts have suggested that data security could be threatened for companies shifting to remote work – as well as ordinary people who are spending more time online – during the coronavirus outbreak.
    &
    In 2019, seven in ten Americans reported feeling less secure about their personal information than they did five years prior. The survey, conducted between June 3-17, 2019, revealed that 6\% of U.S adults felt their personal information was more secure than five years earlier, while 24\% reported feeling that it was the same. These findings suggest that the majority of Americans feel their personal data is more vulnerable now than it was before, a fear that has been compounded by the coronavirus pandemic and the increased reliance on digital communication.
    \\
    \midrule
    \raisebox{-1\height}{\hspace{0mm}  \includegraphics[width=.4\textwidth,trim={0cm 0cm 0cm 0cm},clip]{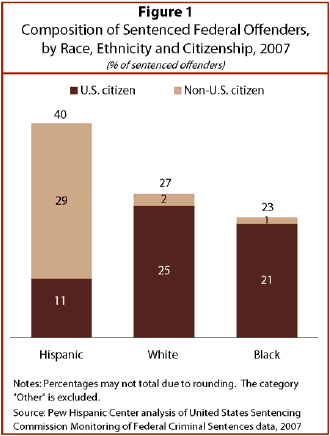}} 
    & Sharp growth in illegal immigration and increased enforcement of immigration laws have dramatically altered the ethnic composition of offenders sentenced in federal courts. In 2007, Latinos accounted for 40\% of all sentenced federal offenders‚Äîmore than triple their share (13\%) of the total U.S. adult population. The share of all sentenced offenders who were Latino in 2007 was up from 24\% in 1991, according to an analysis of data from the United States Sentencing Commission (USSC) by the Pew Hispanic Center, a project of the Pew Research Center.
    & 
    In 2007, 40\% of all federal offenders sentenced in the U.S. were Latino, according to an analysis of United States Sentencing Commission data by the Pew Hispanic Center. This was more than triple the share of Latinos (13\%) in the total U.S. adult population. The share of Latino offenders sentenced in federal courts had risen from 24\% in 1991. 
    \\
    \midrule
    \raisebox{-1\height}{\hspace{0mm}  \includegraphics[width=.4\textwidth,trim={0cm 0cm 0cm 0cm},clip]{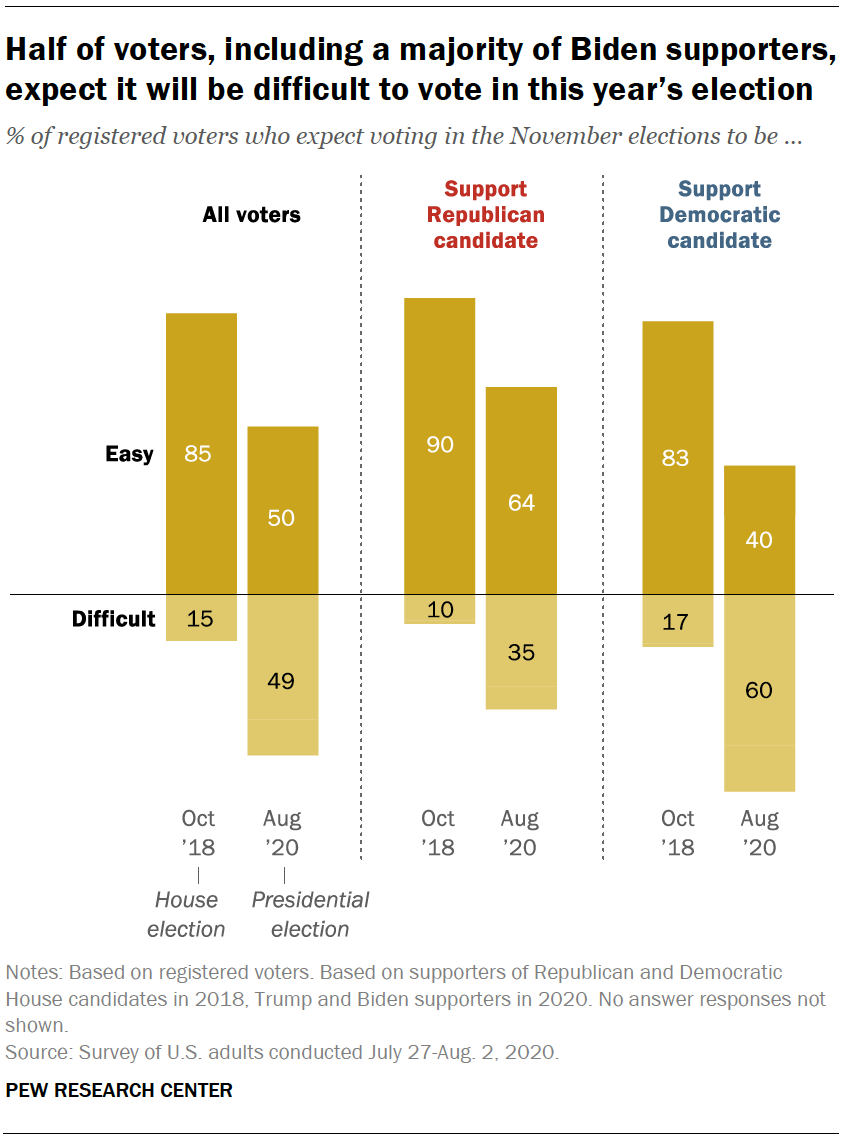}}
    & Far more voters expect voting to be difficult this year than in 2018, according to a survey of registered voters conducted July 27 to Aug. 2. Around half of registered voters (49\%) said they expect voting to be very or somewhat difficult this year, while the other half (50\%) expect it to be very or somewhat easy. That represents a major change from October 2018, when just 15\% of voters said they expected voting in that year‚Äôs midterms to be difficult and 85\% said they expected it to be easy.
    & 
    This year, a majority of voters are expecting it to be difficult to vote in the November elections. According to a survey of registered voters conducted July 27 to Aug. 2, 49\% of registered voters said they expect voting to be very or somewhat difficult this year, while the other half (50\%) expect it to be very or somewhat easy. This is a stark contrast to the survey results from October 2018, when only 15\% of voters said they expected voting in that year‚Äôs midterms to be difficult and 85\% said they expected it to be easy. These results suggest that many Americans are feeling more uncertain and anxious about voting this year than in previous years.
    \\
    \midrule
    \midrule
    \raisebox{-1\height}{\hspace{0mm}  \includegraphics[width=.4\textwidth,trim={0cm 0cm 0cm 0cm},clip]{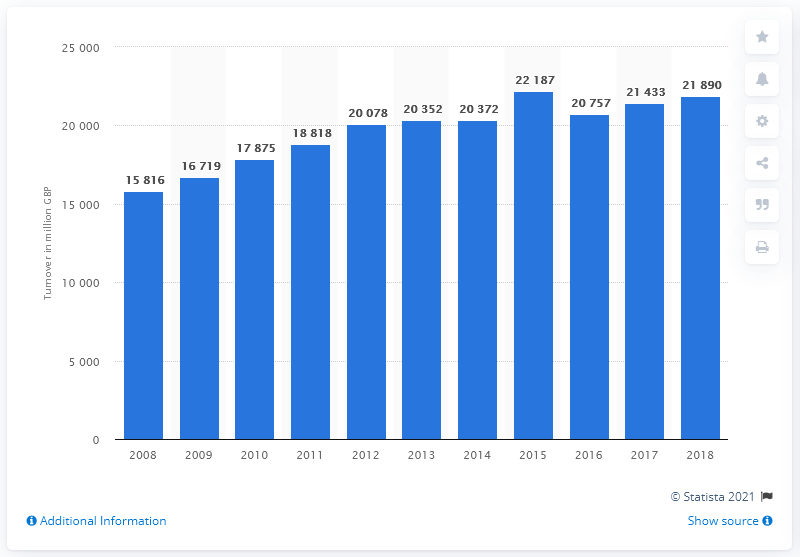}}
    & This statistic shows the total annual turnover of the retail sale of bakery products and cereals ( including rice and pasta products ) in the United Kingdom from 2008 to 2018 . Turnover has grown steadily , reaching a peak of approximately 22 billion British pounds in 2015 before declining the following year . By 2018 turnover had almost completely recovered and amounted to 21.9 billion pounds . 
    & 
    This statistic shows the total annual turnover of the retail sale of bakery products and cereals (including rice and pasta products) in the United Kingdom from 2008 to 2018. Turnover has grown steadily, with a peak of approximately 22 billion British pounds in 2015 before declining the following year. By 2018, turnover had almost completely recovered and amounted to 21.9 billion pounds.
    \\
    \midrule
    \raisebox{-1\height}{\hspace{0mm}  \includegraphics[width=.4\textwidth,trim={0cm 0cm 0cm 0cm},clip]{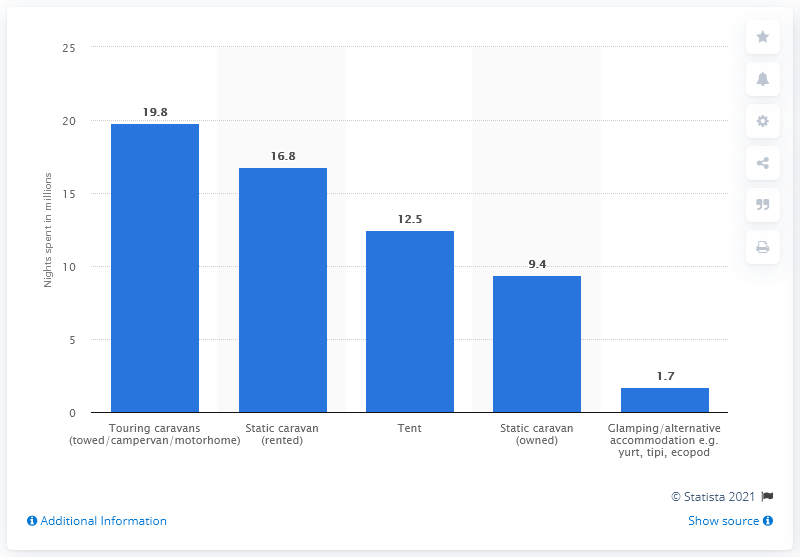}} 
    & This statistic shows the number of nights spent camping or in caravans in Great Britain in 2019 , by type . Almost 20 million nights were spent by domestic tourists in touring caravans in Great Britain , while 12.5 million nights were spent camping in tents . 
    &
    This statistic shows the number of nights spent camping or in caravans in Great Britain in 2019, by type. Nearly 20 million nights were spent by domestic tourists in touring caravans in Great Britain, while 12.5 million nights were spent camping in tents. 
    \\
    \midrule
    \raisebox{-1\height}{\hspace{0mm}  \includegraphics[width=.4\textwidth,trim={0cm 0cm 0cm 0cm},clip]{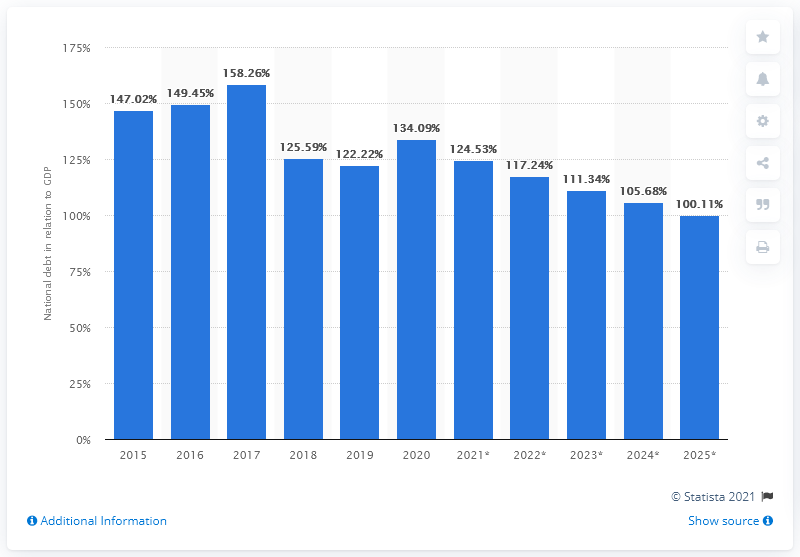}} & 
    This statistic shows the national debt of Barbados from 2015 to 2020 in relation to the gross domestic product ( GDP ) , with projections up until 2025 . The figures refer to the whole country and include the debts of the state , the communities , the municipalities and the social insurances . In 2020 , the national debt of Barbados amounted to approximately 134.09 percent of the GDP .
    & 
    This statistic shows the national debt of Barbados from 2015 to 2020 in relation to the gross domestic product (GDP), with projections up until 2025. The figures refer to the whole country and include the debts of the state, the communities, the municipalities and the social insurances. In 2020, the national debt of Barbados amounted to approximately 134.09 percent of the GDP.\\
    \bottomrule
    \end{tabular}}
  \caption{\small{For each demonstration in our prompt, we ask the language model to generate its expected output, given its input. We then compare the generated output from the model with its expected output.
  }}
  \label{tab:chart2text-demonstrations-consistency}
\end{table*}

\subsection{Analysis of Visual Data Table Representation} \label{appdx:analysis-visual-information}

\begin{table}
\centering
\scalebox{.55}{
\begin{tabular}{lccl}
\hline
Model & Aug. (\%) & Human (\%) & Avg. (\%) \\ 
\hline
InstructGPT + CCR & - & 30.86 & - \\
\model{} & - & 38.42 & -\\
\hline
InstructGPT + CCR + \emph{ground\_truth} colors & - & 60.95 & -\\
 \hline
\end{tabular}
}\label{tab:effect-of-color}
\caption{Effect of color information on color-related QA pairs (203 cases).}
\end{table}

This section is to answer the research question: \textbf{"What is the maximal improvement we can obtain by integrating visual data tables instead of traditional data tables?"} To answer this question, we manually label the colors for a total of 203 color-related QA pairs in the human test set of ChartQA \cite{masry-etal-2022-chartqa}. We then examine the performance of InstructGPT model on these QA pairs in three different settings: (1) with ground-truth visual data tables, (2) with original data tables, (3) with our generated visual data tables from VDTG module. The experimental results are shown in \Cref{tab:effect-of-color}. 
We observe that with the ground-truth colors, the performance of the model is improved significantly, nearly twice compared to the case without them. This observation not only verifies the effectiveness of our proposed visual data table representation but also reveals a huge room for further improvements.

\subsection{Illustrative Cases for Proposed Properties of LCQA and CS Prompts} \label{appdx:proposed-properties-illustrative}

\Cref{tab:illustrative-proposed-properties-lfqa} shows the illustrative of the variants of our LCQA prompt which do not satisfy property \emph{(ii), (iii)} in \Cref{sec:method}. We derive 2 main observations. First, not guaranteeing \emph{(ii)} makes the answers commonly lack findings for \emph{Discovery} queries (first 2 samples). Second, the demonstrations that do not ensure \emph{(iii)} commonly have very long answers. By using these demonstrations not satisfying \emph{(ii), (iii)} in the prompt, we observe that the model commonly generates very long answers which makes them likely to suffer from factual errors (last 2 examples).

\begin{table*}[t!]
  \centering
  \small
  \scalebox{.65}{
    \begin{tabular}{p{9cm}|p{1.5cm}|p{3.5cm}|p{3.5cm}|p{4.5cm}}
    \toprule
    Chart Image & Input & Gold Answer & \model{} & Ablation Study Model \\
    \midrule
    \raisebox{-1\height}{\hspace{0mm}  \includegraphics[width=.45\textwidth,trim={0cm 0cm 0cm 0cm},clip]{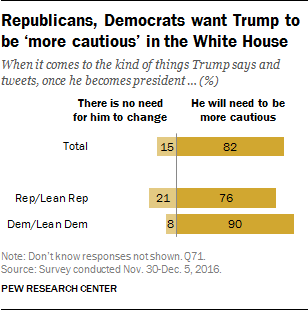}} & Question: How was Trump's online posture seen when he was becoming president?   
    &  In December , a majority of the public ( 82 \% ) also said Trump should be more cautious in “ the kinds of things he says and tweets ” once he becomes president . Just 15 \% said there is no need for him to change .
    &  
     When it comes to the kind of things Trump says and tweets, once he becomes president , 82 \% of Americans believe he will need to be more cautious , including 90 \% of Democrats and 76 \% of Republicans . Only 15 \% of Americans believe there is no need for Trump to change his online posture. &  Most people, including both Republicans and Democrats, believe that Trump needs to be more cautious when it comes to what he says and tweets once he becomes President. 82 \% of the total population, 76 \% of Republicans and Lean Rep and 90 \% of Democrats and Lean Dem think Trump should be more cautious.
     \\
     \midrule
     \raisebox{-1\height}{\hspace{0mm}  \includegraphics[width=.45\textwidth,trim={0cm 0cm 0cm 0cm},clip]{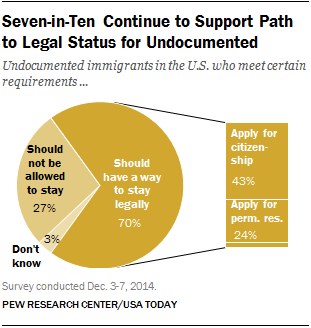}}  & How does public feel about supporting undocumented immigrants? 
     & 70 \% say there should be a way for undocumented immigrants to stay in the country legally , if they meet certain requirements . Most of those who support legal status think there should be a way for unauthorized immigrants to become citizens ( 43 \% overall ) , while 24 \% say the path should only include permanent residency . 
     & A large majority of the public (70\%) support a path to legal status for undocumented immigrants in the US who meet certain requirements. This includes 43\% who support a path to citizenship and 24\% who support a path to permanent residency. Only 27\% of people oppose a path to legal status for undocumented immigrants and 3\% are not sure. 
     &  Seven-in-ten Americans ( 70 \% ) say undocumented immigrants currently living in the US should be allowed to stay in the country legally , if they meet certain requirements , while 27 \% say they should not be allowed to stay . \\
    \midrule
    \midrule
     \raisebox{-1\height}{\hspace{0mm}  \includegraphics[width=.45\textwidth,trim={0cm 0cm 0cm 0cm},clip]{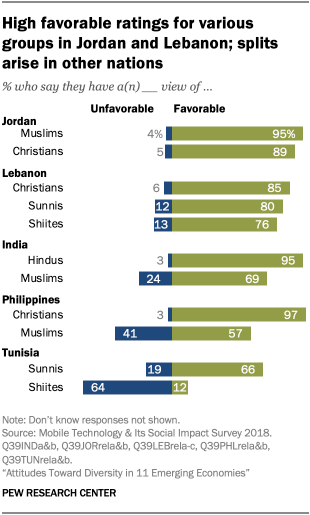}} 
  & What is the favourable ratings of Philippines given by Christians and Muslims ? 
     & in Christian - majority Philippines , favorable views of Christians were nearly universal ( 97 \% ) while just 57 \% said the same of Muslims in their country .
     &
     The favorable ratings for Christians and Muslims in the Philippines are 97 \% and 57 \% respectively.
     & A large majority of the public (70\%) support a path to legal status for undocumented immigrants in the US who meet certain requirements. This includes 43\% who support a path to citizenship and 24\% who support a path to permanent residency. Only 27\% of people oppose a path to legal status for undocumented immigrants and 3\% are not sure. \\
     \midrule
    \raisebox{-1\height}{\hspace{0mm}  \includegraphics[width=.45\textwidth,trim={0cm 0cm 0cm 0cm},clip]{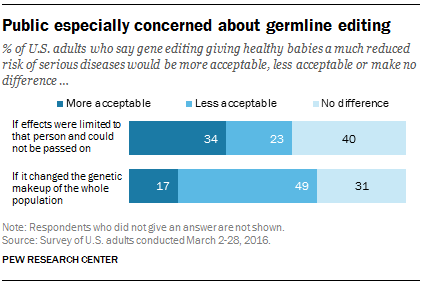}} 
  & What do Americans think about gene editing? & Roughly half of adults ( 49 \% ) said gene editing would be less acceptable to them if the effects “ changed the genetic makeup of the whole population , ” versus just 17 \% who said it would be more acceptable . When asked about an alternate scenario in which the effects of gene editing are limited to a single person , more Americans said it would be more acceptable than less acceptable ( 34 \% to 23 \% ) .
    &  A majority of Americans ( 34 \% ) say that gene editing to give healthy babies a much reduced risk of serious diseases would be more acceptable , while 23 \% say it would be less acceptable and 40 \% say it would make no difference . However , when asked about gene editing that would change the genetic makeup of the whole population , only 17 \% say it would be more acceptable , while 49 \% say it would be less acceptable and 31 \% say it would make no difference. &  Americans are more concerned than accepting when it comes to gene editing . According to a survey conducted in March 2016 by the Pew Research Center , 34 \% of U.S. adults say that gene editing that would give healthy babies a much reduced risk of serious diseases would be more acceptable, while 40 \% say it would make no difference and 23 \% say it would be less acceptable. When it comes to gene editing that would change the genetic makeup of the whole population, only 17 \% say it would be more acceptable, 49 \% say it would be less acceptable and 31 \% say it would make no difference.\\
    \bottomrule
    \end{tabular}}
  \vspace{-2mm}
  \caption{\small{Samples to illustrate the outputs of \model{} and its variant without \emph{($L_2$)} (first 2 rows) and without \emph{($L_3$)} (last 2 rows) for LCQA task.}}
  \label{tab:illustrative-proposed-properties-lfqa}
\end{table*}

\begin{table*}[t!]
  \centering
  \small
  \scalebox{.65}{
    \begin{tabular}{p{9cm}|p{3.5cm}|p{3.5cm}|p{3.5cm}}
    \toprule
    Chart Image & Gold Answer & \model{} & Ablation Study Model \\
    \midrule
    \raisebox{-1\height}{\hspace{0mm}  \includegraphics[width=.3\textwidth,trim={0cm 3.2cm 0cm 0cm},clip]{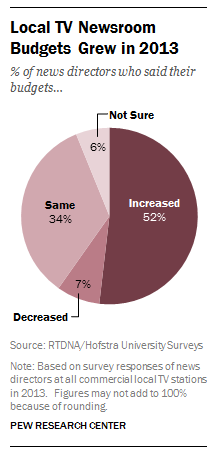}} & In the vast majority of local TV newsrooms, budgets either grew or held steady in 2013. About half of the news directors surveyed (52\%) said they increased their news budgets last year, compared with 49\% the year before and 38\% in 2011. But there were differences by market size. Indeed, 54\% of the mid-sized market stations (51-100) said they increased their budgets compared with 41\% whose budgets were cut or stayed the same. But in the biggest 25 TV markets, the percentage of stations increasing their budgets in 2013 (48\%) was identical to the percentage that cut or did not increase their spending. 
    &  In 2013, local TV newsroom budgets experienced an overall increase. According to a survey of news directors from all commercial local TV stations, 52\% of respondents said their budgets had increased, 34\% said their budgets remained the same, and 6\% were not sure. Just 7\% of news directors reported that their budgets had decreased. These results indicate that local TV newsroom budgets were largely stable or increased in 2013.
    & According to a survey of news directors at all commercial local stations conducted by RTDNA/Hofstra University, 52\% of news directors said their budgets had increased, 34
    \\
    \midrule
    \raisebox{-1\height}{\hspace{0mm}  \includegraphics[width=.3\textwidth,trim={0cm 1.5cm 0cm 0cm},clip]{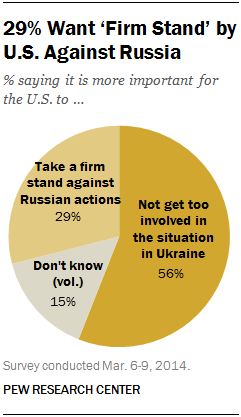}}  
    & As Russian troops remain in Ukraine’s Crimea region and Crimea’s Parliament has set up a secession vote, Americans prefer the U.S. to not get too involved in the situation. By a roughly two-to-one margin (56\% vs. 29\%), the public says it is more important for the U.S. to not get involved in the situation with Russia and Ukraine than to take a firm stand against Russian actions.  
    &  
    In March 2014, Americans were divided on how the U.S. should respond to Russian actions in Ukraine. 29\% believed it was more important for the U.S. to take a firm stand against Russian actions, while 56\% said it was more important not to get too involved in the situation in Ukraine. 15\% said they did not know or gave no opinion.  
    &   
    In 2014, 29\% of Americans said it was more important for the United States to take a firm stand against Russian actions than to not get too involved in the situation in Ukraine. This sentiment was shared among 56\% of those surveyed who chose not to answer. Only 15\% of respondents said that it was more important for the United States not to get too involved in the situation in Ukraine. The survey was conducted by the Pew Research Center from March 6-9, 2014. \\
    \midrule
    \midrule
     \raisebox{-1\height}{\hspace{0mm}  \includegraphics[width=.4\textwidth,trim={0cm 0cm 0cm 0cm},clip]{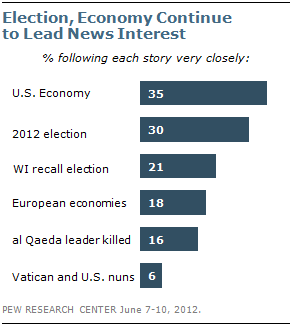}} 
     & 
     The public continued to track news about the economy and the presidential election, while paying less attention to another important political story – Wisconsin Gov. Scott Walker’s victory in a hard-fought recall election. The latest weekly News Interest Index, conducted June 7-10 among 1,000 adults by the Pew Research Center for the People \& the Press, finds that 35\% say they followed reports about U.S. economic conditions very closely while 30\% tracked news about the presidential election very closely. The public expressed less interest in news about the Wisconsin recall election (21\% very closely), European economic problems (18\%) and the CIA killing of an al Qaeda leader in a drone strike in Pakistan (16\%). 
     &  
     In June 2012, 35\% of Americans reported they were following news about the U.S. economy very closely. This was the leading story in terms of interest, with 30\% of Americans saying they were following it very closely. This was followed by the 2012 election, with 21\% of Americans following it very closely. 18\% of Americans were following news about European economies very closely, 16\% were following news about Al Qaeda leader killed, and 15\% were following news about the Vatican and U.S nuns. These figures were reported in a survey conducted by the Pew Research Center from June 7-10, 2012.
    & In June 2012, Americans followed the U.S. economy the closest, with 35\% of adults saying they followed the story “very closely.” The 2012 presidential election and the Wisconsin recall election were the next most closely followed stories, with 30\% and 21\% of adults, respectively, following these stories “very closely.” Europeans economies and the death of al Qaeda leader Osama bin Laden rounded out the top five stories, with 18\% and 16\% of adults saying they followed those stories “very closely.” The Vatican and U.S. nuns dispute was the least followed story, with just 6\% of adults following the story “very closely.” These results are based on a June 7-10, 2012 survey conducted by the Pew Research Center. \\
     \midrule
    \raisebox{-1\height}{\hspace{0mm}  \includegraphics[width=.4\textwidth,trim={0cm 0cm 0cm 0cm},clip]{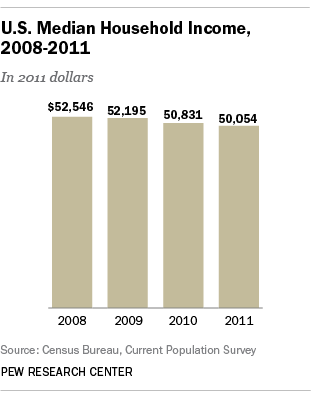}} & As for median household income, the most recent data we have is for 2011 (the Census Bureau will release 2012 numbers next week), but they show that after factoring in inflation, the median U.S. household income shrank 4.7 percent between 2008 and 2011. Little surprise, then, that more than half (54\%) of the people in our survey said household incomes have hardly recovered since the recession, and a scant 2\% said they’ve fully recovered. 
     & 
     From 2008 to 2011, median household income in the United States decreased slightly, according to data from the U.S. Census Bureau's Current Population Survey. In 2011 dollars, median household income was \$52,546 in 2008, \$51,195 in 2009, \$50,831 in 2010, and \$50,054 in 2011. This represents a drop of 4.2\% over the four-year period. &  The median household income in the United States decreased from 2008 to 2011, according to the U.S. Census Bureau's Current Population Survey. In 2008, the median household income was \$52,546, while in 2011, it had fallen to \$50,054. This represented a decrease of 4.1\% in real terms, or a decrease of \$2,492 in inflation-adjusted dollars. The economic recession that began in 2008 was the main factor behind the decrease in median household income during this period. \\
    \bottomrule
    \end{tabular}}
  \vspace{-2mm}
  \caption{\small{Samples to illustrate the outputs of \model{} and its variant without \emph{($S_1$)} (first 2 rows) and without \emph{($S_3$)} (last 2 rows) for CS task.}}
  \label{tab:illustrative-proposed-properties-lfqa}
\end{table*}

\subsection{Error Analysis Case Studies}

\begin{table*}[t!]
  \centering
  \small
  \scalebox{.63}{
    \begin{tabular}{p{9cm}|p{2.5cm}|p{2.5cm}|p{2.5cm}|p{2.5cm}}
    \toprule
    Chart Image & Question & PromptChart & InstructGPT & Gold Answer \\
    \midrule
    \raisebox{-1\height}{\hspace{0mm}  \includegraphics[width=.45\textwidth,trim={0cm 1.5cm 0cm 0cm},clip]{emnlp2020-templates/case-studies/36995393008758.png}}  & How many color bars are there in the graph? & There are \textcolor{ao(english)}{4} color bars in the graph. The answer is \textcolor{ao(english)}{4}. & There is only \textcolor{red}{one} color bar in the graph. The answer is \textcolor{red}{1}. & 4\\
    \midrule
        \raisebox{-1\height}{\hspace{0mm}  \includegraphics[width=.3\textwidth,trim={0cm 3cm 0cm 0cm},clip]{emnlp2020-templates/case-studies/3728.png}} & What is the color of "More"? & The color of the pie represented "More" is \textcolor{ao(english)}{Gray}. The answer is  \textcolor{ao(english)}{Gray}. & The color of the pie represented "More" is \textcolor{red}{Orange}. The answer is \textcolor{red}{Orange}. & {Gray}\\
     \midrule
     \raisebox{-1\height}{\hspace{0mm}  \includegraphics[width=.35\textwidth,trim={0cm 2cm 0cm 0cm},clip]{emnlp2020-templates/case-studies/12051.png}} & What is the colour of India in the graph? & The colour of India in the graph is \textcolor{ao(english)}{orange}. The answer is \textcolor{ao(english)}{orange}. &  The colour of India in the graph is \textcolor{red}{not given in the data table}. Therefore, the answer is \textcolor{red}{not given}. & {Orange} \\
    \midrule
    \raisebox{-1\height}{\hspace{0mm}  \includegraphics[width=.35\textwidth,trim={0cm 2.5cm 0cm 0cm},clip]{emnlp2020-templates/case-studies/7780.png}} & Is the color of right-side bars lightest? &  The color of right-side bars is  \textcolor{ao(english)}{light brown}. The lightest color of the bars is  \textcolor{ao(english)}{light brown}, the answer is  \textcolor{ao(english)}{Yes}. & The right-side bars are \textcolor{red}{Dem/Lean Dem and Rep/Lean Rep}. The color of \textcolor{red}{Dem/Lean Dem bar is lightest}. The answer is  \textcolor{ao(english)}{Yes}. & Yes \\
    \midrule
    \midrule
    \raisebox{-1\height}{\hspace{0mm}  \includegraphics[width=.5\textwidth,trim={0cm 1.5cm 0cm 0cm},clip]{emnlp2020-templates/case-studies/01001540004402.png}} & Which color Bar shows the greatest Meat consumer? & The greatest meat consumer is \textcolor{ao(english)}{Finland} with 175.09. The color of Finland is \textcolor{ao(english)}{Teal}. So the answer is \textcolor{red}{Teal}. & The highest meat consumption is in Finland with 175.09. So the color bar that shows the greatest meat consumer is the bar for \textcolor{ao(english)}{Finland}. The answer is \textcolor{ao(english)}{Finland}. & {Finland}\\
    \midrule
    \raisebox{-1\height}{\hspace{0mm}  \includegraphics[width=.5\textwidth,trim={0cm 1.5cm 0cm 0cm},clip]{emnlp2020-templates/case-studies/67940673002511.png}} & What country had the highest yearly cost of closing global poverty in 2002? & The highest yearly cost of closing global poverty in 2002 is \textcolor{red}{Comoros} \textcolor{ao(english)}{(gray)} with 16.87. The answer is \textcolor{red}{Comoros}. & The highest yearly cost of closing global poverty in 2002 is Botswana with 144.7. The answer is \textcolor{ao(english)}{Botswana} & {Botswana}\\
    \bottomrule
    \end{tabular}}
  \vspace{-2mm}
  \caption{\small{The first four cases show PomptChart answers correctly by visual information whilst InstructGPT answers wrongly. The last two cases illustrate that visual information sometimes harms the model's performance. Both models are run in the few-shot setting with our proposed FCQA prompt.}}
  \label{tab:chatqa-cases}
\end{table*}

\Cref{tab:error-analysis} shows our case studies as discussed in \Cref{sec:error-analysis}.

\begin{table*}[t!]
  \centering
  \small
  \scalebox{.6}{
    \begin{tabular}{p{6.5cm}|p{5.5cm}|p{13cm}}
    \toprule
    \textbf{Question 1:} What is the average of all three bars?
    
    \raisebox{-1\height}{\hspace{0mm}  \includegraphics[width=.4\textwidth,trim={0cm 0cm 0cm 0cm},clip]{emnlp2020-templates/imgs/23355169003897.png}}

    \textbf{Generated Answer 1:} The average of the three bars is \textcolor{red}{(1.5 + 1.34 + 1.2) / 3 = 1.32}. The answer is 1.32.

    \textbf{Gold Answer 1:} 1.346
     & 
    \textbf{Question 2:} Is the average percentage of all three bars greater than 6\%? 

    \raisebox{-1\height}{\hspace{0mm}  \includegraphics[width=.35\textwidth,trim={0cm 0cm 0cm .5cm},clip]{emnlp2020-templates/imgs/26535003011580.png}}

    \textbf{Generated Answer 2:} The percentage of all three bars is \textcolor{ao(english)}{( 7.19 + 6.85 + 6.2 ) / 3 = 6.7466666667\%}. The answer is \textcolor{red}{No, the average is not greater than 6\%}

    \textbf{Gold Answer 2:} Yes
    & 
     \raisebox{-1\height}{\hspace{0mm}  \includegraphics[width=.43\textwidth,trim={0cm 0cm 0cm .5cm},clip]{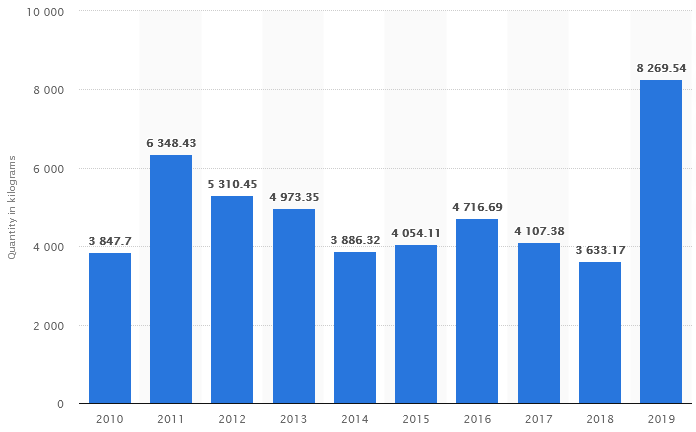}}  

     \textbf{Generated Summary 3:} This statistic shows the quantity of cocaine seized in Italy from 2010 to 2019 . \textcolor{ao(english)}{In 2019 , 8269.54} kilograms of cocaine were seized in Italy , a significant increase from the \textcolor{ao(english)}{3847.7 kilograms seized in 2010} . The \textcolor{red}{largest amount of cocaine seized during this period was in 2011 , with 6348.43 kilograms} .

     \textbf{Gold Summary 3:} The statistic shows the total quantity of cocaine seized in Italy from 2010 to 2019 . In 2019 , 8.3 thousand kilograms of cocaine were seized in the country , the largest amount registered between 2010 and 2019 .\\
     
    \midrule
    \midrule  
    
    \textbf{Question 4:} What are the opinions of Independents on Obama?
    \raisebox{-1\height}{\hspace{0mm}  \includegraphics[width=.2\textwidth,trim={0cm 1.5cm 0cm 0cm},clip]{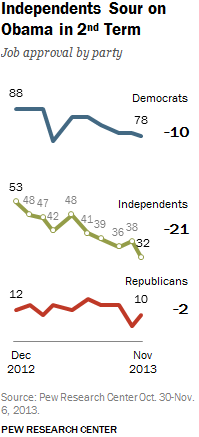}} 

    \textbf{Generated Answer 4:} During Obama's second term , there has been a significant drop in support among Independents . \textcolor{red}{In December 2012 , 48 \%} of Independents approved of Obama's job performance , compared to only \textcolor{ao(english)}{32 \% in November 2013} , a decrease of 16 percentage points . By comparison , Democrats' approval of Obama's job performance only \textcolor{red}{decreased by 10 percentage points ( 78 \% to 88 \% )} and Republicans' approval \textcolor{red}{decreased by only 2 percentage points ( 10 \% to 12 \% )} .

    \textbf{Gold Answer 4:} Obama's approval ratings among Independents declined from 48 \% in December 2012 to 36 \% in November 2013, a drop of 21 points, which is larger than 10 point decline among Democrats and 7 point decline among Republicans.
     & 
     \textbf{Question 5:} What are the views of the general public on LGBT figures?
    
    \raisebox{-1\height}{\hspace{0mm}  \includegraphics[width=.25\textwidth,trim={0cm 0cm 0cm .5cm},clip]{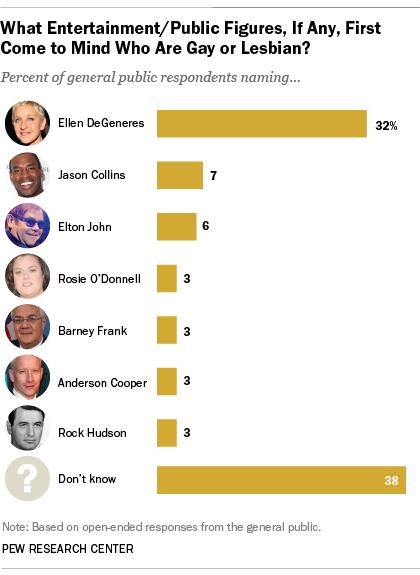}}  

    \textbf{Generated Answer 5:} Among the general public, 32 \% of respondents named Ellen DeGeneres and 38 \% named Elton John as the first LGBT artist or public figure that comes to mind. Other names mentioned included Rosie O'Donnell \textcolor{red}{( 11 \% )} , Jason Collins \textcolor{red}{( 8 \% )} , Barney Frank \textcolor{red}{( 4 \% )} , Anderson Cooper \textcolor{red}{( 4 \% )} and Rock Hudson \textcolor{red}{( 4 \% )} . A total of \textcolor{ao(english)}{38 \%} of respondents said they didn't know.

    \textbf{Gold Answer 5:} Jason Collins was the second - most frequently mentioned gay or lesbian public figure . 38 \% said they could n’t name any gay or lesbian entertainers or public figures . 
     &
     \raisebox{-1\height}{\hspace{0mm}  \includegraphics[width=.45\textwidth,trim={0cm 1.5cm 0cm 0cm},clip]{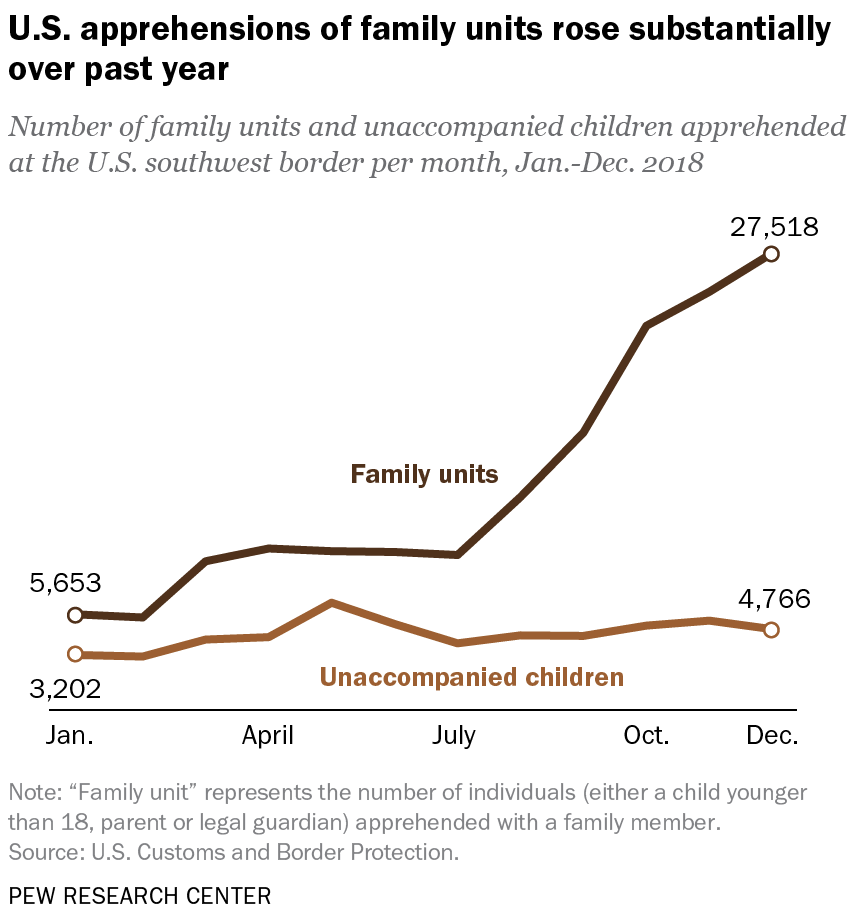}} 

     \textbf{Generated Summary 6:} In 2018, the number of family units and unaccompanied children apprehended at the U.S. southwest border rose substantially, reaching a peak of 27,518 family units and 5,653 unaccompanied children in December. The number of apprehensions for both groups increased steadily throughout the year, starting at 4,766 family units and 3,202 unaccompanied children in January. The increase in family unit apprehensions is especially notable, as the number of family units apprehended in 2018 was nearly six times higher than the number apprehended in 2017. \textcolor{red}{Experts suggest that the rise in apprehensions is due to a combination of factors, including increasing violence in Central American countries and more expansive U.S. immigration policies}.

     \textbf{Gold Summary 6:} In the months leading up to the shutdown, there was a large increase in the number of people in family units apprehended at the border. Monthly family apprehensions subsequently hit new highs each month from September through December, according to data going back to 2012. There were nearly 17,000 family member apprehensions in September, more than 23,000 in October, about 25,000 in November and a record of more than 27,000 in December.
     
     \\
    \bottomrule
    \end{tabular}}
  \vspace{-2mm}
  \caption{\small{Case studies for error analysis. \textcolor{red}{Red} texts are errors and \textcolor{ao(english)}{green} ones are correct generations.}
  }
  \label{tab:error-analysis}
\end{table*}

\subsection{Benchmarks Statistics} \label{appdix:benchmark-statistics}
We present the statistics of the benchmarks we use in \Cref{tab:dataset-summarization}. ChartQA focuses on logical and visual reasoning questions, while OpenCQA targets open-ended questions with explanatory answers and Chart-to-Text is a benchmark for chart summarization task.

    